\documentclass[conference]{IEEEtran}
\IEEEoverridecommandlockouts
\usepackage[utf8]{inputenc}
\usepackage{cite}
\usepackage{amsmath,amssymb,amsfonts}
\usepackage{algorithmic}
\usepackage{graphicx}
\usepackage{subcaption}
\usepackage{siunitx}
\usepackage{textcomp}
\usepackage{xcolor}
\usepackage{fancyhdr}
\usepackage[ruled,vlined]{algorithm2e}
\def\BibTeX{{\rm B\kern-.05em{\sc i\kern-.025em b}\kern-.08em
    T\kern-.1667em\lower.7ex\hbox{E}\kern-.125emX}}
\begin{document}

\title{HAMLET - A Learning Curve-Enabled Multi-Armed Bandit for Algorithm Selection}

\author{
    \IEEEauthorblockN{Mischa Schmidt, Julia Gastinger, S\'{e}bastien Nicolas, Anett Sch\"ulke }
    \IEEEauthorblockA{NEC Laboratories Europe GmbH, Kurf\"ursten-Anlage 36, 69115 Heidelberg, Germany
    \\\{FirstName.LastName\}@neclab.eu}    

}

\maketitle

\begin{abstract}
Automated algorithm selection and hyperparameter tuning facilitates the application of machine learning. Traditional multi-armed bandit strategies look to the history of observed rewards to identify the most promising arms for optimizing expected total reward in the long run. When considering limited time budgets and computational resources, this backward view of rewards is inappropriate as the bandit should look into the future for anticipating the highest final reward at the end of a specified time budget. This work addresses that insight by introducing HAMLET, which extends the bandit approach with learning curve extrapolation and computation time-awareness for selecting among a set of machine learning algorithms. Results show that the HAMLET Variants 1-3 exhibit equal or better performance than other bandit-based algorithm selection strategies in experiments with recorded hyperparameter tuning traces for the majority of considered time budgets. The best performing HAMLET Variant 3 combines learning curve extrapolation with the well-known upper confidence bound exploration bonus. That variant performs better than all non-HAMLET policies with statistical significance at the 95\% level for 1,485 runs. 

 \end{abstract}


\section{Introduction}\label{sec:intro}
When applying machine learning, one crucial decision to be taken is to choose a learning algorithm (denoted \textit{base learner}) among the plethora of algorithms available. Each machine learning algorithm comes with a different set of hyperparameters that can be optimized to maximize the algorithm's performance concerning an application-specific error metric for a given dataset. Besides, different feature preprocessing algorithms and feature selection techniques - each with their set of hyperparameters - can be combined into a machine learning pipeline to improve the base learner's performance. 

Automated machine learning (AutoML) addresses the automation of selecting base learners and preprocessors as well as tuning the associated hyperparameters. Hence, AutoML lowers the workload of expert data scientists by automating many of the decision steps. Further, AutoML provides a structured approach to identify well-performing base learner configurations. AutoML typically outperforms manual tuning or grid search heuristics \cite{wistuba2016hyperparameter} (and references therein). Moreover, AutoML allows nonexperts to leverage machine learning efficiently, e.g.~in the biomedical domain \cite{luo2016review}.

This work focuses on selecting the base learner to be applied to a dataset. Specifically, this work models the iterative approach to selecting the base learner and the optimization of its hyperparameters as a hierarchical problem. A multi-armed bandit focuses on selecting the base learner, and a specialized component (the \textit{tuner}) is responsible for tuning that respective base learner's hyperparameters. This approach is easily extensible with base learners by integrating them as additional arms. %(and their respective tuner [TODO @Mischa i would remove this in the brackets, ok?])  

In realistic settings, AutoML faces a limitation of resources in terms of computational power and time budget for solving the machine learning problem. This work studies the extreme case of a single CPU available for solving a machine learning task within a strict wallclock time budget. In this setting, the traditional multi-armed bandit approach is not optimal because it requires to observe a complete function evaluation. In other words, it requires training a parametrized base leaner on the dataset before updating the associated arm's statistics. Additionally, most multi-armed bandit algorithms assume stationary reward distributions - which in AutoML is not true as tuning algorithms usually increase their performance over time. Finally, the typical AutoML problem strives to get the maximum possible performance, not to maximize the average sum of rewards over repeated trials. For these reasons, this work modifies the multi-armed bandit approach by accounting for time explicitly and learning the different arms' learning curves. The bandit then solves the algorithm selection problem by extrapolating the curves to the end of the time budget with consideration of the budget already spent on each arm. 

We refer to the approach presented in this work as HAMLET - Hierarchical Automated Machine LEarning with Time-awareness due its hierarchical decision making and its ability to account for the progress of time. We summarize the hypothesis underlying HAMLET as follows:
\begin{quote}
The combination of learning curve extrapolation and accounting for computation time improves the performance of multi-armed bandits in algorithm selection problems.
\end{quote}

Our empirical evaluation uses the 99 traces for six different base learners from \cite{8851978}. The evaluation shows that even a simple approach to fit learning curves provides gains for tight time budgets. Overall, the best performing HAMLET Variant 3 achieves with 95\% confidence a better performance than all non-HAMLET bandits used in the experiments.

The remainder of this work is structured as follows. Section \ref{sec:related} summarizes relevant related work and shows where this work differs. Section \ref{sec:methods} introduces the notation, the methods, and the experimental protocol to verify HAMLET's effects. Section \ref{sec:experiments} documents the experiments' results that are discussed in Section \ref{sec:discussion}. Finally, Section \ref{sec:conclusion} concludes this work by summarizing the main findings and providing an outlook on future work.

\section{Related Work}\label{sec:related}
\subsection{Exploration and Exploitation for Multi-Armed Bandits}
%The basic form of the multi-armed bandit problem is, e.g.~described by \cite{sutton2018reinforcement}: 
In the basic form of the multi-armed bandit problem, an agent is faced repeatedly with a choice among $k$ different actions. After each choice, the agent receives a numerical reward from a stationary probability distribution that depends on the selected action. The objective is to maximize the expected total reward over some time period.  %Through repeated action selections, the agent can maximize the collected rewards by concentrating on the best arms.
At any time step, the agent may act greedily and select one of the actions with the highest estimated value (denoted exploitation). On the other hand, by selecting a non-greedy action, the agent can improve its estimate of the chosen action’s value (denoted exploration).   Agents need to explore, because even though the greedy actions are those that look best at present, some of the other actions may actually be better \cite{sutton2018reinforcement}.  
%A common exploration technique is called `optimistic initialization'. That technique initializes action values to high values. As a consequence, the agent chooses these initially attractive actions, observes lower rewards, and subsequently lowers its action value estimates. Non-visited actions remain at their high initial values, which attracts the agent to explore these actions in future. 

A simple exploration technique called  `$\epsilon$-greedy' behaves most of the time greedily, but with small probability $\epsilon$ selects randomly from among all the actions with equal probability. An advantage is that as the number of time steps increases, the bandit will sample every action an infinite number of times. Therefore, the bandit's action value estimates will converge to the accurate values. %Commonly, $\epsilon$-greedy action selection forces the non-greedy actions to be tried, but indiscriminately.
Another technique denoted \textit{decaying $\epsilon$} initializes $\epsilon$ high and decreases it (and thus, the rate of exploration) over time \cite{sutton2018reinforcement}. 

%Another effective way of selecting among the possible actions is the Upper Confidence Bound (UCB) method. 
In contrast to random exploration, the Upper Confidence Bound (UCB) method selects actions according to their potential for being optimal.
 UCB does so by taking into account both their respective value estimates and the uncertainties in those estimates. UCB will select actions with lower value estimates, and actions that have already been selected frequently with decreasing frequency over time.
%UCB bandits select actions based on the upper bound of what is reasonable to assume as the highest possible true value for each action. 
%Each time an action is selected, the epistemic uncertainty in the action value estimate should decrease. %On the other hand, each time another action is selected, the epistemic uncertainty increases.
One difficulty of UCB bandits is  in  dealing  with non-stationary problems \cite{sutton2018reinforcement}. 
%... Some explanation of the UCB concept, e.g. from Sutton and Barto or so.
% ATM: BestK velocity and rewards use UCB

The multi-armed bandit problem presented in this work differs from the original problem as rewards are not stationary. When performing algorithm selection, the rewards should increase as more time is spent on the arm, while the rate of improvement is unknown. The objective is not to maximize the total reward, but to find the single best reward.

\subsection{Multi-Armed Bandits for Algorithm Selection and Hyperparameter Tuning}
ATM \cite{Swearingen2017} is a distributed, collaborative, scalable AutoML system, which incorporates algorithm selection and hyperparameter tuning. ATM approaches AutoML by iterating two steps: hyperpartition selection followed by hyperparameter tuning.  A hyperpartition includes one specific base learner, as well as its categorical hyperparameters. ATM models each hyperpartiton selection as a multi-armed bandit problem. ATM supports three bandit algorithms: the standard UCB-based algorithm `UCB1', and two variants designed to handle drifting rewards as encountered in the AutoML setting. The variants compute the value estimates for selecting the actions either based on the velocity or the average of the best K rewards observed (denoted \textit{BestK-Velocity} and \textit{BestK-Rewards}, respectively).  %Once a hyperpartition has been chosen, the remaining unspecified parameters can be selected from a vector space using, for example, Bayesian Optimization. 
The Machine Learning Bazaar \cite{Smith2019} framework for developing automated machine learning software systems extends the work of ATM and incorporates the same bandit structures. HAMLET differs from both. First, it does not choose between hyperpartitions, but solely between base learners, i.e.~it does not select categorical hyperparameters. Second, it uses a novel bandit algorithm, which fits a simple model of the learning curve to observed rewards, but selects the action based on an extrapolation of the learning curve to find the highest possible reward given a time budget. Third, ATM and the Machine Learning Bazaar update the action value statistics based on completed function evaluations, i.e.~a base learner's test performance after training it on the dataset. HAMLET updates training statistics in configurable time intervals. Even if a base learner's tuner did not manage to find better models in a recent time interval, HAMLET tracks (the lack of) progress of the tuner's learning curve, allowing it to switch computing resource assignments based on extrapolating learning curves. This work uses the bandits of \cite{Swearingen2017} as baselines to compare HAMLET to.

Hyperband \cite{Li2017HyperbandBC} is a bandit-based early-stopping method for sampled parametrizations of base learners. It incorporates a bandit that deals with the fact that arms in hyperparameter optimization might improve when given more training time. Hyberband builds on the concept of successive halving: it runs a set of parametrized base learners for a specific budget, evaluates their performances, and stops the worse performing half of the set. 
When presented with a larger set of possible parametrizations of base learners, Hyperband stops parametrizations that do not appear promising and assigns successively more computational resources to the promising ones that remain. 
%
%Hyperband \cite{Li2017HyperbandBC} is a multi-armed bandit-based early-stopping method for sampled parametrizations of base learners. This way, when presented a larger set of possible parametrizations of base learners for a dataset, Hyperband iteratively stops parametrizations that do not appear promising and consequently assigns more computational resources to more promising ones. Thus, Hyperband requires that the base learner algorithm provides performance statistics during training, and that it benefits when it is continued in training. For example deep learning falls into that category - more epochs typically increases the neural network performance. 
% HAMLET differs in that it assigns computational resources for a time interval to a tuner and does not require base learner statistics being available at intermediate time steps. (? @ MISCHA verstehe nicht was du meinst. die machen das doch auch in iterations)
HAMLET differs in that it assigns computational resources based on predicted performance and not observed performance. In HAMLET, the bandit is used to decide which algorithm to run, and not which hyperparameter setting to run.
Also, the approach to assign budget is different. Hyperband applies the concept of a geometric search to assign increasing portions of the overall budget to a decreasing number of base learner parametrizations. In contrast, HAMLET continues a chosen tuner for a configured time interval. %After the interval, the tuner reports updates for the best found models, if any, and HAMLET updates the respective tuner's learning curve.

\subsection{Learning Curve Extrapolation}
As outlined in \cite{Domhan2015}, the term learning curve is used to describe (1) the performance of an iterative machine learning algorithm as a function of its training time or number of iterations and (2) the performance of a machine learning algorithm as a function of the size of the dataset it has available for training. For the AutoML challenge addressed by HAMLET, we focus on extrapolating type (1) learning curves.

%Targeting hyperparameter optimization of Deep Neural Networks (DNN), \cite{Domhan2015} utilizes a probabilistic model to extrapolate the performance from the first part of a learning curve for a hyperparameter configuration. 
Given the first part of a learning curve for a given hyperparameter configuration of a Deep Neural Network (DNN), \cite{Domhan2015} utilizes a probabilistic model to extrapolate its performance. For this, \cite{Domhan2015} fits a set of parametric functions for each hyperparameter configuration and combines them into a single model by weighted linear combination. A Markov Chain Monte Carlo method yields probabilistic extrapolations of the learning curve, which allows to terminate runs with non-promising hyperparameter settings early automatically.
Relying on a Bayesian Neural Network (BNN) in combination with the parametric functions presented by \cite{Domhan2015}, \cite{Klein2017LearningCP} samples promising candidates to apply Hyperband \cite{Li2017HyperbandBC} to. Predicting the model parameters of parametric functions as well as the mixing weights with the BNN enables transferring knowledge of previously observed learning curves. However, that implies that previous learning curve information is needed to pre-train the BNN for good performance.
%
%(\cite{Klein2017LearningCP} used Bayesian Neural Networks to learn the parameters of a set of saturating functions and their combinations to predict deep networks performances and use that to sample promising candidates for Hyperband \cite{Li2017HyperbandBC}. This work differs as we use a less sophisticated learning curve function to demonstrate the general nature of moving from a backward to a forward-looking Bandit.  Moreover, the learning curves are of the tuners, not individual parametrized base learner instances. )

Freeze-Thaw optimization \cite{Swersky2014} is a Gaussian Process-based Bayesian optimization technique for hyperparameter search. The method includes a learning curve model based on exponential decay and a positive definite covariance kernel to model the iterative optimization curves. %Gaussian Process-based Bayesian optimization technique for hyperparameter search. The method includes a learning curve model based on a parametric exponential decay model. They use a positive definite covariance kernel to model the iterative optimization curves. 
The Freeze-Thaw method maintains a set of partially completed but not actively trained models and uses the learning curve model for deciding in each iteration which ones to `thaw', i.e.~to continue training. 
% present a Gaussian Process based optimaization technique. In this, they use a positive definite covariance kernel to model iterative optimization curves. In this context they create a parametric exponential decay model. The learning curves are used to tmeporally pause training of models with certain hyperparameter connfigurations and resume at a later stage.

For speeding up hyperparameter optimization, \cite{Chand2017} proposes a regression-based model for learning curve extrapolation.  
%\cite{Chand2017} propose a regression-based extrapolation model for the extrapolation of learning curves to speed up hyperparameter optimization. 
The technique relies on trajectories from previous builds to make predictions of new builds, where a `build' refers to a training run with a specific base learner parametrization. Therefore, \cite{Chand2017} transforms data from previous builds and adds a noise term to match the current build and to extrapolate its performance in order to identify and stop hyperparameter configurations early. 

HAMLET differs from previous work as, first, we use a less sophisticated learning curve function to demonstrate the general nature of benefits derived from moving from a backward to a forward-looking multi-armed bandit for algorithm selection. Second, this work aspires to provide a general approach, not limited to a specific type of base learner such as DNN \cite{Domhan2015}, \cite{Klein2017LearningCP}.  Third, to test our presented hypothesis, this work does not rely on previous learning curves \cite{Klein2017LearningCP}, \cite{Chand2017},  but only uses information from the current AutoML problem. While a transfer of information from previous learning curves appears beneficial, this work investigates whether HAMLET improves algorithm selection performance due to a simple learning curve extrapolation, not due to a transfer and reuse of previous information. Finally, HAMLET extrapolates learning curves for the performance of base learners' tuners and not individual hyperparameter configurations. %We extrapolate the learning curves with the goal of giving more budget to more promising algorithms (context of algorithm selection) instead of early stopping of less promising hyperparameter configurations for already chosen algorithms.

\section{Methods}\label{sec:methods}
\subsection{HAMLET}
\subsubsection{Symbols and Notation}
Tab.~\ref{tab:notation} summarizes the symbols used in this work.

\begin{table}[t]
\caption{Symbols and notation used in this work.}\label{tab:notation}
\begin{tabular}{r|l}
Symbol & Description \\\hline 
$B$ & overall Budget given \\
$B_{rem}$ & remaining time budget \\

$\epsilon_1$ & HAMLET Variant 1: chance to pick the tuner with the  \\& second-highest learning curve \\
$\epsilon_2$ & HAMLET Variant 1: chance to pick a tuner at random \\
$\epsilon_t$ & HAMLET Variant 2: chance to pick a tuner at random, \\& time dependent \\
$\textrm{I}$ & number of arms \\

$LC^i$ & Learning Curve Function for arm $i$, e.g.~using Eq.~\ref{eq:atan}\\

$\mathbf{r}$ & vector of predicted reward of all arms when budget runs out \\
$\hat{r}^i$, $\hat{\mathbf{r}} $ & $r^i$, resp.~$\mathbf{r}$,  after applying UCB bonus (Eq.~\ref{eq:ucb})\\
$\rho$ & HAMLET Variant 3: UCB exploration bonus scaling factor \\

$\Delta t$ & time interval for HAMLET's main loop in algorithm 1\\
$t_x^i$ & execution time of arm $i$ spent until now\\
$r^i$ & predicted reward of arm $i$ when budget runs out \\

$\left[\mathbf{x}^i, \mathbf{y}^i \right]$ & observed learning curve values for arm $i$ \\
$\mathbf{x}^i$ & training time to reach $\mathbf{y}^i$  for arm $i$ \\
$\mathbf{y}^i$ & accuracy values for arm $i$ \\
$\mathbf{\hat{y}}^i$ & predicted accuracy values for arm $i$ \\
\end{tabular}
\end{table}

\subsubsection{Multi-Armed Bandit with Learning Curve Extrapolation}\label{sec:HAMLET-LCE}
Inspired by \cite{Swearingen2017} and \cite{Smith2019}, we model the AutoML algorithm selection problem as a multi-armed bandit problem, where each arm represents a hyperparameter tuner responsible for one specific base learner. 
Each iteration, HAMLET chooses which action to take, i.e.~selects the arm to pull, based on extrapolations of the arms' learning curves as described below in Section \ref{sec:LCE} and outlined in Alg.~\ref{alg:HAMLET_main} and \ref{alg:LCE}. After deciding on the action, the bandit continues the execution of the corresponding hyperparameter tuner for a pre-configured time interval $\Delta t$. When the interval elapses, the arm's execution pauses. This work assumes that the arm's execution resumes in later iterations without loss of information.  During the execution of the arm, the bandit receives all monotonically increasing accuracy values reached in that time interval as well as information about when the arm reached these values (i.e.~the updated observed learning curve). When the arm's tuner did not find new monotonically increasing accuracy values in that time interval, this information is also incorporated. The bandit then fits a parametric curve to match the learning curve and reduces the remaining budget by $\Delta t$. Subsequently, HAMLET continues to the next iteration. 

When HAMLET faces a new AutoML problem, it tries arms in a Round Robin fashion for a predefined amount of time to collect enough values to model learning curves for each arm. 

\subsubsection{Learning Curve Extrapolation}\label{sec:LCE}
\begin{figure}[!t]
\centering
	\includegraphics[width=0.8\columnwidth]{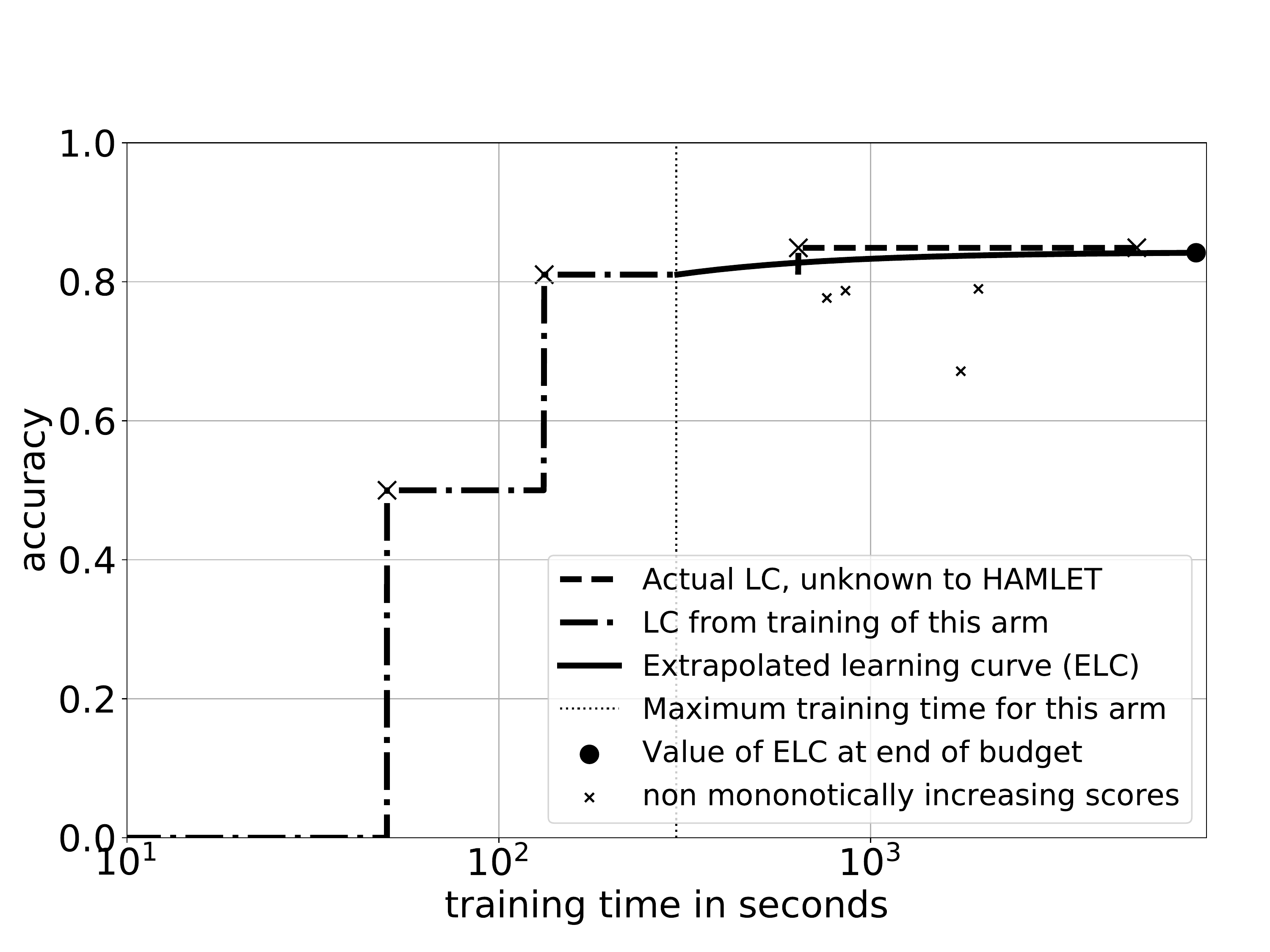}
\caption{Exemplary graph showing how a learning curve (LC) is extrapolated based on observed learning curve values. %Also we show the ground truth learning curve, which is unknown to the bandit for comparison as well as other accuracy scores found by the tuner, which are not monotonically increasing and are thus not used for learning curve extrapolation.\textbf{TODO im text beschreiben}.
}\label{fig:lc}
\end{figure}

%In HAMLET, the learning curve is defined as a monotonically increasing function defined by the maximum accuracies found over training time.
This work defines the learning curve as a monotonically increasing function defined by the maximum accuracies found over training time. 
Fig.~\ref{fig:lc} depicts an example explaining HAMLET's approach to learning curve extrapolation. In this work, the term training time includes the time spent on executing the tuner for identifying parametrizations of the base learner and training the parametrized base learner on the dataset.
In Fig.~\ref{fig:lc}, an arm has been running for $500$ seconds (marked by a vertical line). From the accuracy values found until this point, we extract the monotonically increasing values and use those to fit a learning curve approximation. That approximation serves to predict future accuracy values. % From the accuracy scores found by this tuner, only monotonically increasing values are used for the learning curve, other scores are ignored. 
For comparison, Fig.~\ref{fig:lc} also shows the arm's actual future learning curve, i.e.~the ground truth unknown to the HAMLET bandit after executing the arm for $500$ seconds. 
The goal in AutoML is to find, in a specified time budget $B$, the model leading to the best test score. In this work that score is given by the maximum accuracy value across all tuners, i.e.~arms $i \in \{1,...,\textrm{I}\}$, where each arm has received an amount of computational resources, in this case training time $t_x^i$: $\max_{t_x^i}{y^i_{t_x^i}} \textrm{, s.t. } \sum_{i=1}^{\textrm{I}} {t_x^i} \leq B$.
 %\begin{equation}\label{eq:goal1}
%\max_{i}{y^i} \textrm{, s.t. } \sum_{i=1}^{\textrm{I}} {t_x^i} \leq B.
%\end{equation}
%In our case, the accuracy $y$ is the maximum accuracy value across all tuners, i.e.~arms $i \in \{1,...,\textrm{I}\}$, where each arm has received an amount of computational resources, in this case training time $t_x^i$, with $\sum_{i=1}^{\textrm{I}} t_x^i \leq B$. The number of possible arms is defined by $\textrm{I}$.
% \begin{equation}\label{eq:goal1}
%a = \textrm{max } a^i(t_x^i) \textrm{, where } \sum_{i=1}^{\textrm{I}} t_x^i \leq B.
%\end{equation}
%To reach the goal defined in~(\ref{eq:goal1}), 

HAMLET attempts to devote most computational resources to the tuner corresponding to the base learner achieving the highest accuracy, using the specified budget $B$ optimally.   
% \begin{equation}\label{eq:goal2}
%t_x^i = \textrm{argmax } a(i)\textrm{, } i \in{1,...,\textrm{I}}, \sum_{i=1}^{\textrm{I}} t_x^i \leq B
%\end{equation}
%, i.e.~arm $i$, where the number of possible arms is defined by $\textrm{I}$.
% \begin{equation}\label{eq:goal1}
%\textrm{max } t_x^{i_{\textrm{max}}} \textrm{, }   t_x<= B,
%\end{equation}
%where $t_x^{i_{\textrm{max}}}$ is the time devoted to the tuner reaching the highest accuracy, with 
% \begin{equation}\label{eq:goal1}
%i_{\textrm{max}} = \textrm{argmax } a(i)\textrm{, } i \in{1,...,\textrm{I}},
%\end{equation}
%where $a(i)$ is the accuracy to be attained by arm $i$, and 
%where $\textrm{I}$ is the total number of arms. 
As HAMLET iteratively assigns training time intervals to the different arms, it observes a time series of test scores $\left[\mathbf{x}^i, \mathbf{y}^i \right]$, where $x^i \leq t_x^i \textrm{, }\forall x^i \in \mathbf{x}^i$. In any iteration, the arms' future accuracies $\mathbf{y}^i_{x > t_x^i}$ are unknown. Therefore, HAMLET attempts to predict each arm's future accuracy scores $\mathbf{\hat{y}}^i_{x > t_x^i}$ by approximating each arms's learning curve $LC^i$ and extrapolating it. More specifically, HAMLET regresses the arms' observed $\left[\mathbf{x}^i, \mathbf{y}^i \right]$ and extrapolates assuming that all remaining budget $B_{rem}$ was spent exclusively on the corresponding arm $i$: $r^i = \hat{y}_{t_x^i + B_{rem}}^i$. 
%\begin{equation}\label{eq:max}
%r^i = y_{t_x^i + B_{rem}}^i \textrm{, } i \in \{1,...,\textrm{I}\}.
%\end{equation}

To investigate if learning curve extrapolation is a meaningful concept for the algorithm selection problem in constrained computational settings, we use a straightforward parametric function to model the arms' observed accuracies over time. In this work, learning curves are known to be (1) monotonically increasing, (2) saturating functions with (3) values $ \hat{y} \in \left[0, 1\right]$. Because of the similar shape and its compatibility with prerequisites (1)-(3), we choose the arctangent function with four parameters ($a$, $b$, $c$, $d$) to translate, stretch and compress:
 \begin{equation}\label{eq:atan}
%	y = a + b \arctan{c(x + d))}
 \hat{y}_x = a \cdot \arctan{(b(x + c))} + d.
\end{equation}
We use SciPy's \cite{virtanen2020scipy} \texttt{curve fit} function to fit the parameters of the desired curve. 

\subsubsection{HAMLET Variants}
HAMLET faces the same exploration - exploitation dilemma as other multi-armed bandit strategies. This work uses three variants of how HAMLET chooses the arm to run in the next iteration. All variants (see Alg.~\ref{alg:LCE}) base their decision on the $i$-dimensional vector $\mathbf{r}$ containing for each arm $i$ the predicted accuracy $r^i$.  %by extrapolating the learning curve to time  $x=t_x^i + B_{rem}$, $r^i$ . 
%Alg.~\ref{alg:LCE} outlines the variants.

\textit{Variant 1 - Double $\epsilon$-greedy Learning Curve Extrapolation with Fixed $\epsilon_1$ and $\epsilon_2$}: In this approach, HAMLET acts in an $\epsilon$-greedy fashion based on the extrapolation of learning curves. After observing in preliminary experiments that often a subset of the tuners perform much better than the rest, we modified the standard $\epsilon$-greedy bandit as follows. With chance $\epsilon_2$, HAMLET chooses an action at random. With chance $\epsilon_1$, HAMLET chooses the arm with the second-highest predicted accuracy. With chance $1-(\epsilon_1+\epsilon_2)$, HAMLET takes the greedy action, i.e.~$\mathrm{argmax}(\mathbf{r})$. 

\textit{Variant 2 - $\epsilon$-greedy Learning Curve Extrapolation with Decaying $\epsilon$}: In this approach, HAMLET acts in an $\epsilon$-greedy fashion based on the extrapolation of learning curves. The variant starts with $\epsilon_0=1$ %$\epsilon(0)=1$ 
and reduces it in $\frac{B}{\Delta t}$ iterative steps to $\epsilon_B=0$, %to $\epsilon(B)=0$, 
where the notation $\epsilon_t$ %$\epsilon(t)$ 
denotes the stochastic exploration parameter's time dependence. Each iteration, HAMLET chooses an action at random with chance of the current $\epsilon_t$. %$\epsilon(t)$. 
With chance $1-\epsilon_t$, %$1-\epsilon(t)$, 
HAMLET takes the greedy action. 

\textit{Variant 3 - Learning Curve Extrapolation with Exploration Bonus}: This variant adds for each arm a scaled UCB-based exploration bonus \cite{sutton2018reinforcement} to the learning curve predictions to compute the action values: 
\begin{equation} \label{eq:ucb}
%\hat{r^i} = r^i + \rho\sqrt{\frac{2\log{\sum_i{dim(\textbf{x}^i)}}}{\log{ dim(\textbf{x}^i)}}} \text{, } \rho \geq 0
\hat{r}^i = r^i + \rho\sqrt{\frac{2\log{n}}{\log{ n^i}}} \text{, } \rho \geq 0,
\end{equation}
where $n$ is the number of total iterations, $n^i$ is the number of times arm $i$ has been pulled and $\rho$ is the scaling factor of the exploration bonus.
Each iteration, HAMLET chooses the arm with maximum $\hat{r}^i$, i.e.~$\mathrm{argmax}(\mathbf{\hat{ r}})$.

\begin{algorithm}
\small
\SetAlgoLined
%\KwResult{Write here the result }
\KwData{Overall budget $B$}
$B_{rem} = B$

 \While{$B_{rem}$  $> 0$ }{
	\uIf{ First iteration}{
%		Train each arm  $i = 1,...,\mathrm{I}$ once for $\Delta t$ and observe AccuracySteps$\left[\mathbf{x}^i, \mathbf{y}^i \right]$  \;	
	\For{each arm $i = 1,...,\mathrm{I}$ }{
		$\left[\mathbf{x}^i, \mathbf{y}^i \right]$  = TrainAndObserveLC(i, $\Delta t$),\,	 
		where $\left[\mathbf{x}^i, \mathbf{y}^i \right]$ describes the observed learning curve values
	}
	}
	\Else{
		NextArm $=$  MasterChooseNextArm($\mathbf{r}$), \, see Alg.~\ref{alg:LCE}\;
		$\left[\mathbf{x}^i, \mathbf{y}^i \right]$  = TrainAndObserveLC(NextArm, $\Delta t$),\,	 
		where $\left[\mathbf{x}^i, \mathbf{y}^i \right]$ describes the observed learning curve values
	}
	\For{each arm $i = 1,...,\mathrm{I}$ }{
%	$\left[\mathbf{x}^i, \mathbf{y}^i \right]$ = UpdateLC()\;
	$LC^i$ = SciPy.Curve$\_$Fit(Eq.~\ref{eq:atan}, $\left[\mathbf{x}^i, \mathbf{y}^i\right] $)\;
%	$\left[\hat{\mathbf{x}}^i, \hat{\mathbf{y}}^i\right]$ = UpdateExtrapolatedLC($\left[\mathbf{x}^i, \mathbf{y}^i\right] $ )\;
	$r^i$ = $LC^i(t_x^i + B_{rem})$  
%	$r^i$  = PredictValueOfInterest($B_{rem}$ , $t_x^i$, $\left[\hat{\mathbf{x}}^i, \hat{\mathbf{y}}^i\right]$ ) 
	}
	$\mathbf{r}= \left[r^1, r^2, \dots, r^{\mathrm{I}}\right]$\;
	$B_{rem}$  = UpdateBudget() \;
  } 
 \caption{Algorithm selection based on learning curve extrapolation.}\label{alg:HAMLET_main}
\end{algorithm}
\begin{algorithm}
\small
\SetAlgoLined
%\KwResult{Write here the result }
\KwData{ Vector  $\mathbf{r}$ with predicted accuracy for all arms at end of budget}
\#This includes 3 Variants\;
\uIf{Variant 1}{
	with probability $(1-(\epsilon_1+\epsilon_2))$:  $ na =$ argmax$(\mathbf{r})$\;
	with probability $(\epsilon_1)$: $ na =$ $i$, \, if $r^i$ is runner-up in $\mathbf{r}$\; %argmax$(\mathbf{r})$\; 
	with probability $(\epsilon_2)$: $ na =$ random$(1,..,\mathrm{I})$\;   }
\uElseIf{Variant 2}{
	with probability $(1-\epsilon_t):$ $na  =$ argmax$(\mathbf{r})$\;  %#(t)):$ $na  =$ argmax$(\mathbf{r})$\;
	with probability $(\epsilon_t)$: $na  =$ random$(1,..,\mathrm{I})$\;   % (t))$: $na  =$ random$(1,..,\mathrm{I})$\;   
	where $\epsilon_t$  linearly decreases with incr. time $t$\;   %(t)$  linearly decreases with incr. time $t$\;   
}
\Else{	
	calculate $\hat{\mathbf{r}}$ by applying Eq.~\ref{eq:ucb} to all arms\;
	$na =$ argmax $\hat{\mathbf{r}}$\;   
}
 Return Nextarm = $na$\
 \caption{MasterChooseNextArm%, choose next arm based on predicted accuracy at end of budget.
}\label{alg:LCE}
\end{algorithm}

\subsection{Experimental Validation}\label{sec:experiments}
This work  leverages the traces of experiments in \cite{8851978}, which executed hyperparameter tuning for six base learners by an evolutionary strategy. 
Running different algorithm selection policies on the recorded experiment traces allows evaluating different bandit policies based on a common ground truth. While executing the different bandits on recorded traces allows us to skip time-consuming base learner training, we make sure to reflect the time needed to perform the bandits' calculations in the experiment analysis. That enables a fair comparison among the bandits: more expensive bandit strategies have to offset their higher computational costs by selecting better actions.

\subsubsection{Computational Resources and Setup}
%This work leverages the traces of experiments in \cite{8851978}, which 
 \cite{8851978} executed each tuner (and base learner) in a single docker container with only a single CPU core accessible. Parallel execution of different experiments was limited to ensure that a full CPU core was available for each docker container. There was no limit on memory resource availability. This work executes the bandit logic also in docker containers constrained to a single CPU core. We limit the execution of different experiment runs to ensure each docker container has access to one full CPU core.

\subsubsection{Datasets, Base Learners and Hyperparameter Tuners}\label{sec:datasets}
%This work leverages recorded experiment traces of the tuning experiments in \cite{8851978}. 
Relying on the traces from \cite{8851978}, the experiments in this work perform algorithm selection for 49 small\footnote{www.OpenML.org datasets: \{23, 30, 36,  48, 285, 679, 683, 722, 732, 741, 752, 770, 773, 795, 799, 812, 821, 859, 862, 873, 894, 906, 908, 911, 912, 913, 932, 943, 971, 976, 995, 1020, 1038, 1071, 1100, 1115, 1126, 1151, 1154, 1164, 1412, 1452, 1471, 1488, 1500, 1535, 1600, 4135, 40475\}} and 10 bigger classification datasets\footnote{www.OpenML.org datasets: \{46, 184, 293, 389, 554, 	772, 917, 1049, 1120, 1128\} with five repetitions each.}. The datasets have 1,000 - 680,000 samples,  2 - 18 classes, and 3 - 11,000 features. The datasets have no missing feature values and the features are of types numerical or categorical. Tab.~\ref{tab:experiment_settings} documents the budgets used for the experiments with the small (denoted \textit{Experiment 1}) and bigger datasets (\textit{Experiment 2}).
The traces contain six base learners% from \cite{pedregosa2011scikit}%, with hyperparameters tuned by an evolutionary strategy
: k-Nearest Neighbors, linear and kernel SVM, AdaBoost, Random Forest, and Multi-Layer Perceptron.  

\subsubsection{Policies and Parametrizations for Comparative Evaluation}
%Running different algorithm selection policies on the recorded experiment traces allows evaluating different bandit policies based on a ground-truth. 
For HAMLET Variants 1-3, time progresses in intervals of $\Delta t=10s$. This work assumes the capability to freeze and continue the execution of different arms (e.g.~via standard process control mechanisms). 
%For a fair comparison with the other bandit policies, the experiments take the computation time needed for fitting the arms' learning curves into account \textbf{(TODO @ Julia: is this clearer now?)}. @Mischa TODO: I removed it for now but you can of course re-add it.
%This work compares HAMLET Variants 1-3 with a simple Round Robin strategy, a standard UCB1 bandit, BestK-Rewards and BestK-Velocity policies, leveraging the BTB library \cite{Smith2019}. For each parametrizable policy (BestK-Rewards, BestK-Velocity, HAMLET variants 1 and 3) we ran a simple grid search (refer Tab.~\ref{tab:experiment_settings}) to identify the best performing parameter for that policy when considering all datasets and all budgets. This mimics a realistic setting, where the Data Scientist may not know the optimal policy parametrization beforehand and thus parametrizes based on an educated guess. 
This work compares HAMLET Variants 1-3 with a simple Round Robin strategy (\textit{`Round Robin'}), a standard UCB1 bandit (\textit{`UCB'}), BestK-Rewards (\textit{`BestKReward-}$K$', where $K$ refers the parameter choice used) and BestK-Velocity (\textit{`BestKVelocity-}$K$') policies, leveraging \cite{Smith2019}. The HAMLET Variant 1 is presented by \textit{`MasterLC}-$\epsilon_1$-$\epsilon_2$'. HAMLET Variant 2 relates to \textit{`MasterLCDecay'} and \textit{`MasterLC-UCB-$\rho$'} refers to HAMLET Variant 3.  For each parametrizable policy (BestK-Rewards, BestK-Velocity, HAMLET Variants 1 and 3), we ran a simple grid search (refer Tab.~\ref{tab:experiment_settings}) to identify the best performing parameter for that policy when considering all datasets and all budgets. That mimics a realistic setting, where the Data Scientist may not know the optimal policy parametrization beforehand. Thus, she parametrizes based on an educated guess. 
\begin{table}
\caption{Verification experiment parameter sets.}
\label{tab:experiment_settings}
\begin{tabular}{r|l}
Parameter & Parameter Set\\\hline
$\epsilon_1$ & $\{0.01, 0.05, 0.10, 0.20, 0.40, 0.60\}$\\
$\epsilon_2$ & $\{0.00, 0.01, 0.05, 0.10, 0.20, 0.40\}$\\
$\rho$ & $\{0.00, 0.05, 0.10, 0.25, 0.50, 0.75, 1.00\}$\\
$K$ & $\{3, 5, 7, 10, 20, 50, 100\}$\\
B (Exp.~1) [s] & $\{150, 300, 450, 600, 900, 1800, 3600\}$ \\
B (Exp.~2) [s] & $\{900, 1800, 2700, 3600, 7200, 10800, 21600,43200\}$ 
\end{tabular}
\end{table}
\subsubsection{Analysis}
We compare the highest accuracies achieved by each policy parametrization per dataset within a given budget using boxplots to identify the most promising choices of $K$, $\epsilon_1$, $\epsilon_2$, and $\rho$. After identifying each policy's best performing parametrizations across the different budgets, we compare these against each other. Finally, statistical inference yields the intervals of 95\% confidence for the policies' mean ranks in these inter-policy comparisons.  

\section{Experiment Results}\label{sec:experiments}
%\subsection{Experiment 1: Small Datasets}
\begin{figure*}[!t]
%\begin{minipage}{.24\textwidth}
  %\centering
  %\includegraphics[width=\textwidth]{img/olympic/intra/intra_pol_ranks_IQR__MasterLC-0__b600.pdf}
  %\subcaption{\label{fig:b600}Budget 10 minutes. }
%\end{minipage}
%\hfill
\begin{minipage}{.32\textwidth}
	\centering
	\includegraphics[width=\textwidth]{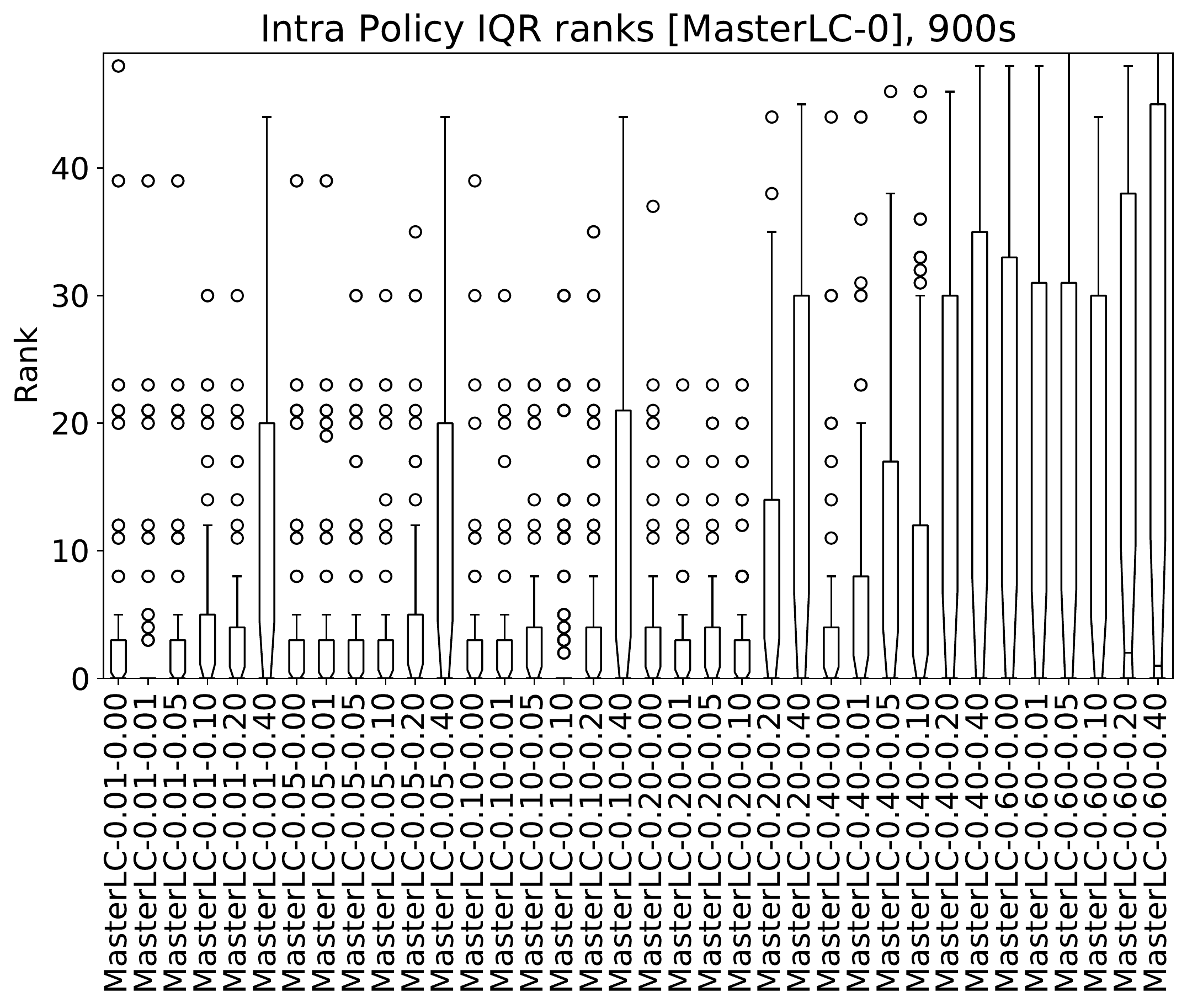}
	\subcaption{\label{fig:b900}Budget 15 minutes.}
\end{minipage}
\hfill
\begin{minipage}{.32\textwidth}
	\centering
	\includegraphics[width=\textwidth]{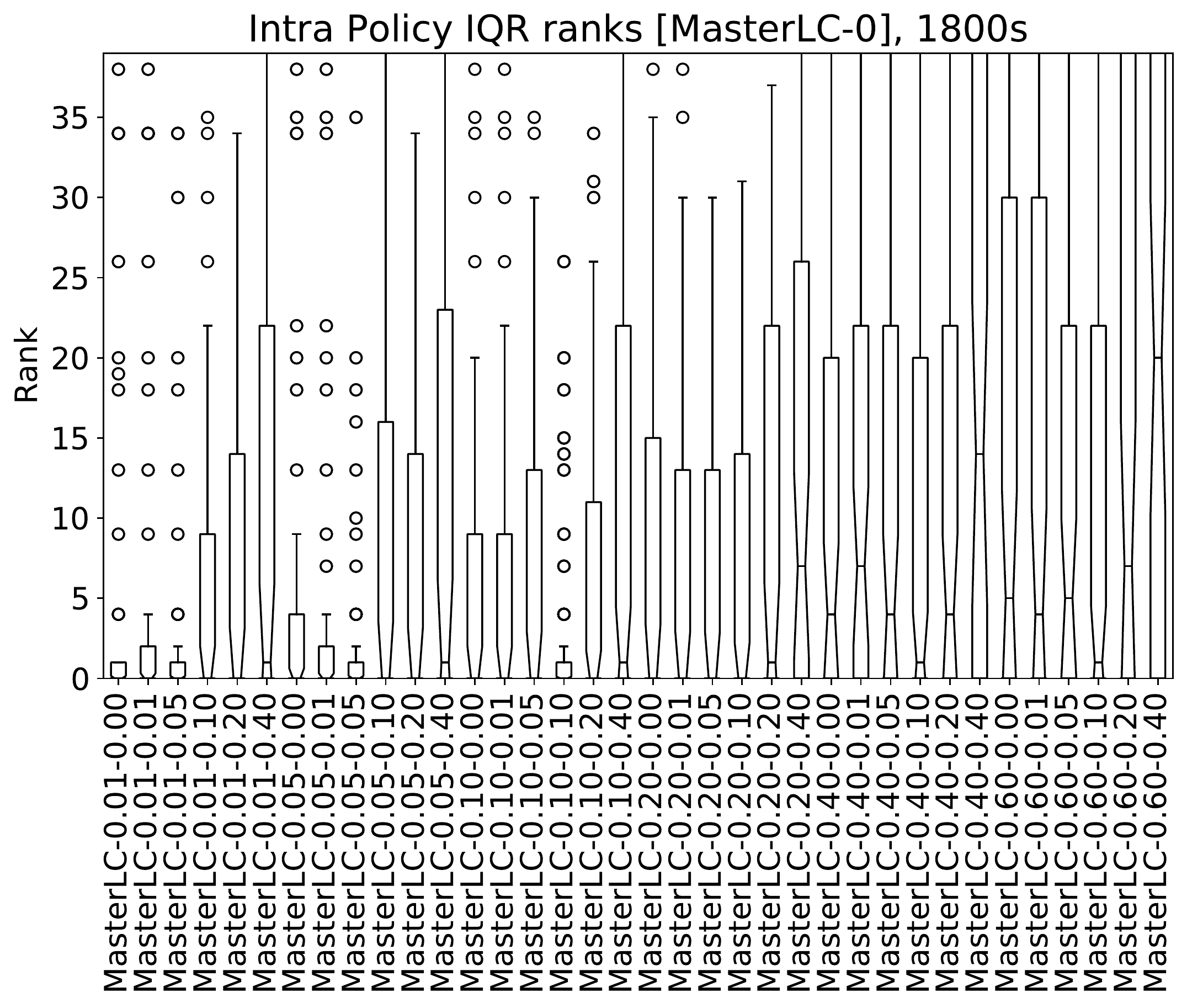}
	\subcaption{Budget 30 minutes.}
\end{minipage}
\hfill
\begin{minipage}{.32\textwidth}
	\centering
	\includegraphics[width=\textwidth]{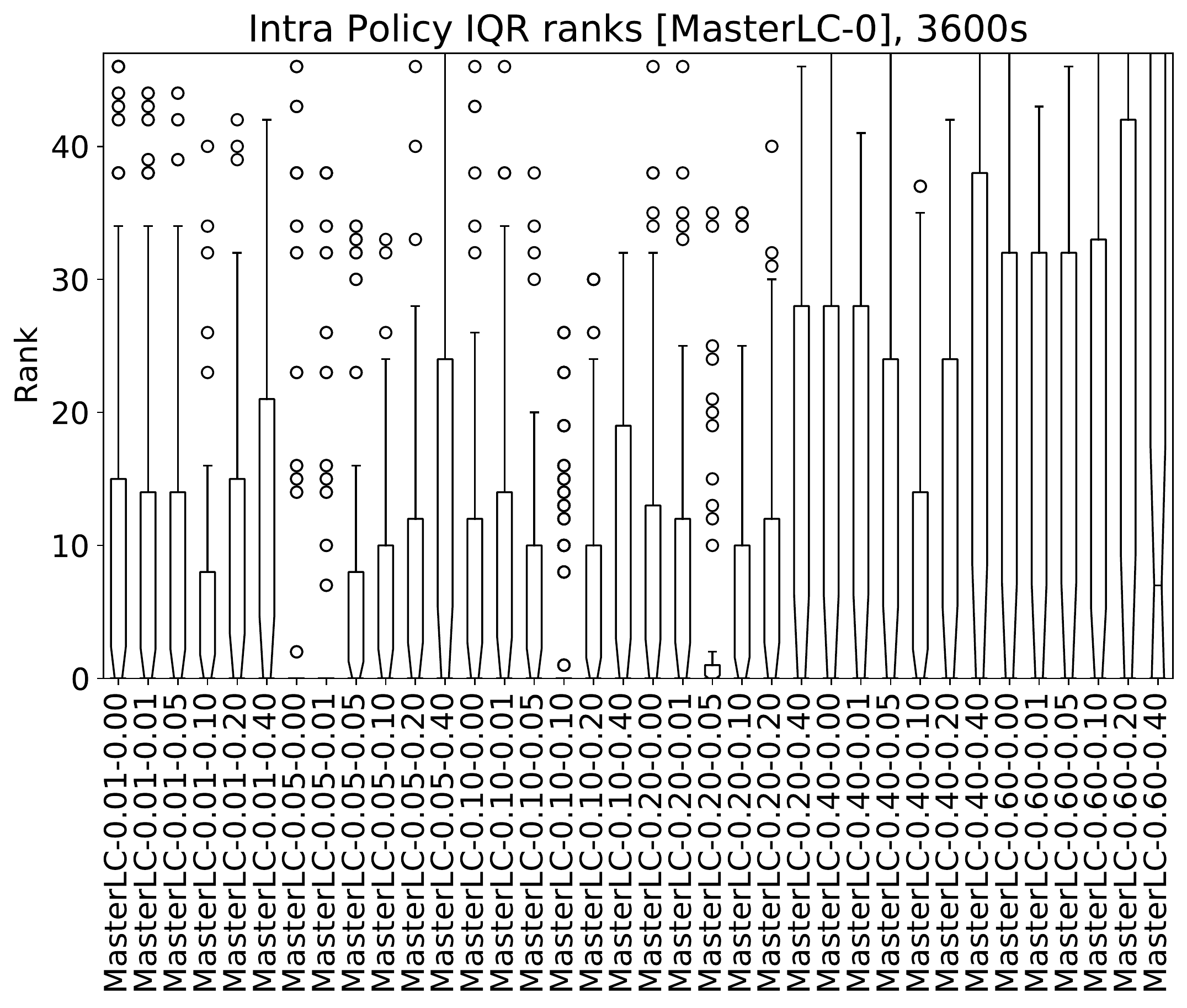}
	\subcaption{Budget 1 hour.}
\end{minipage}

\caption{Experiment 1: Boxplots of HAMLET Variant 1 ranks. With smaller budgets the results do not change qualitatively.}
	\label{fig:exp1_HAMLET-V1}
\end{figure*}

\begin{figure*}[!t]
\begin{minipage}{.22\textwidth}
  \centering
  \includegraphics[width=\textwidth]{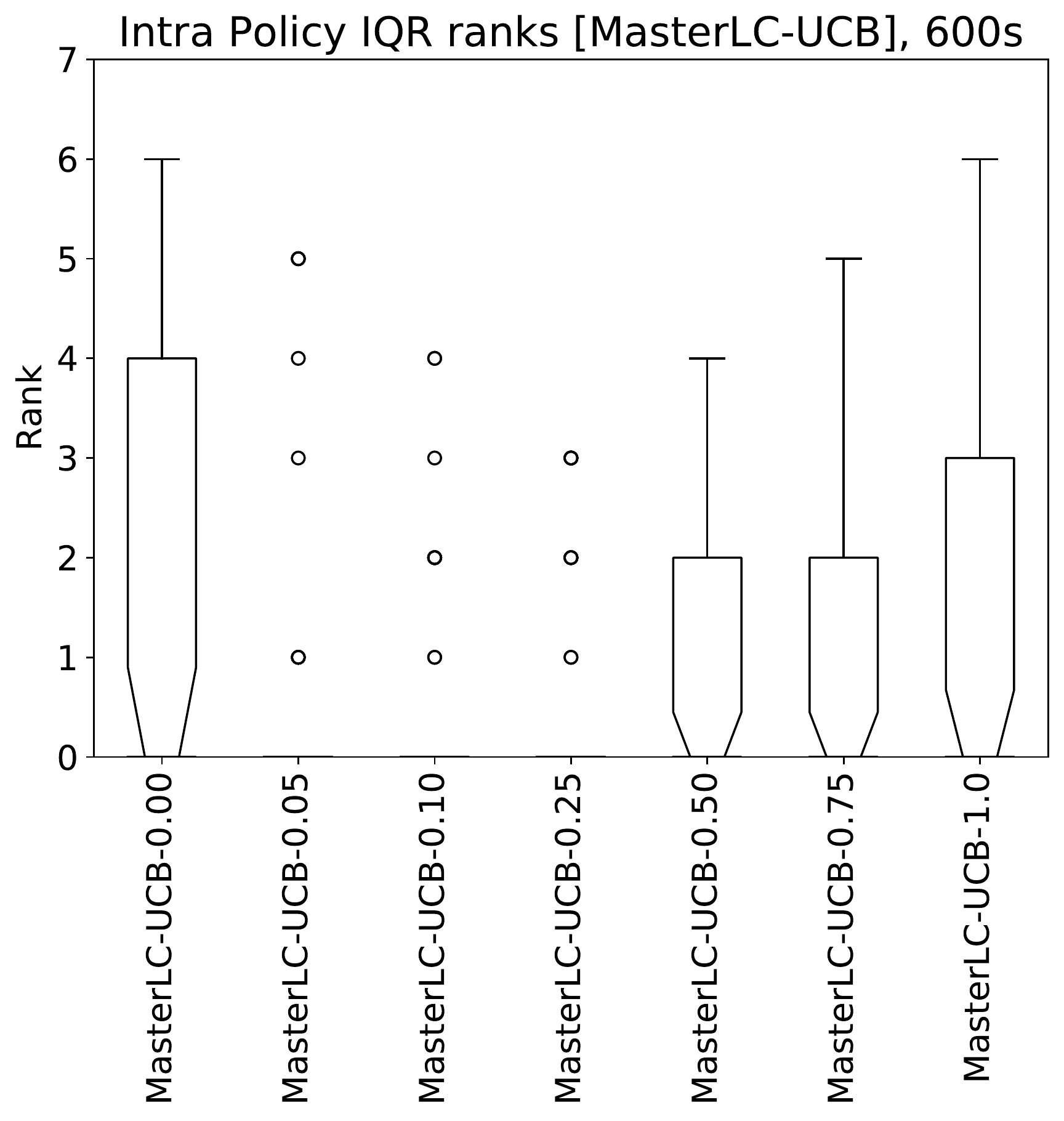}
  \subcaption{\label{fig:b600}Budget 10 minutes. }
\end{minipage}
\hfill
\begin{minipage}{.22\textwidth}
	\centering
	\includegraphics[width=\textwidth]{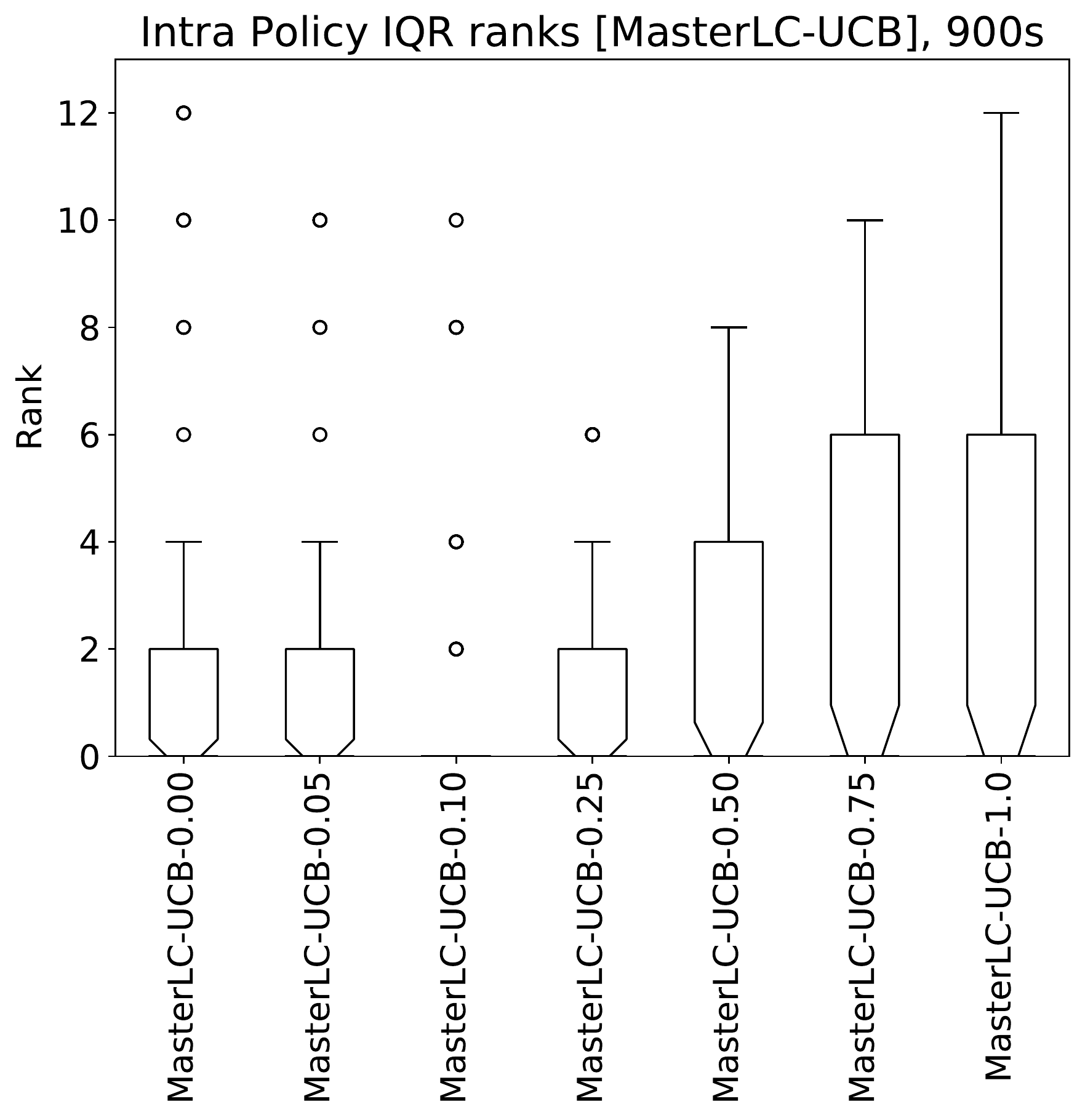}
	\subcaption{\label{fig:b900}Budget 15 minutes.}
\end{minipage}
\hfill
\begin{minipage}{.22\textwidth}
	\centering
	\includegraphics[width=\textwidth]{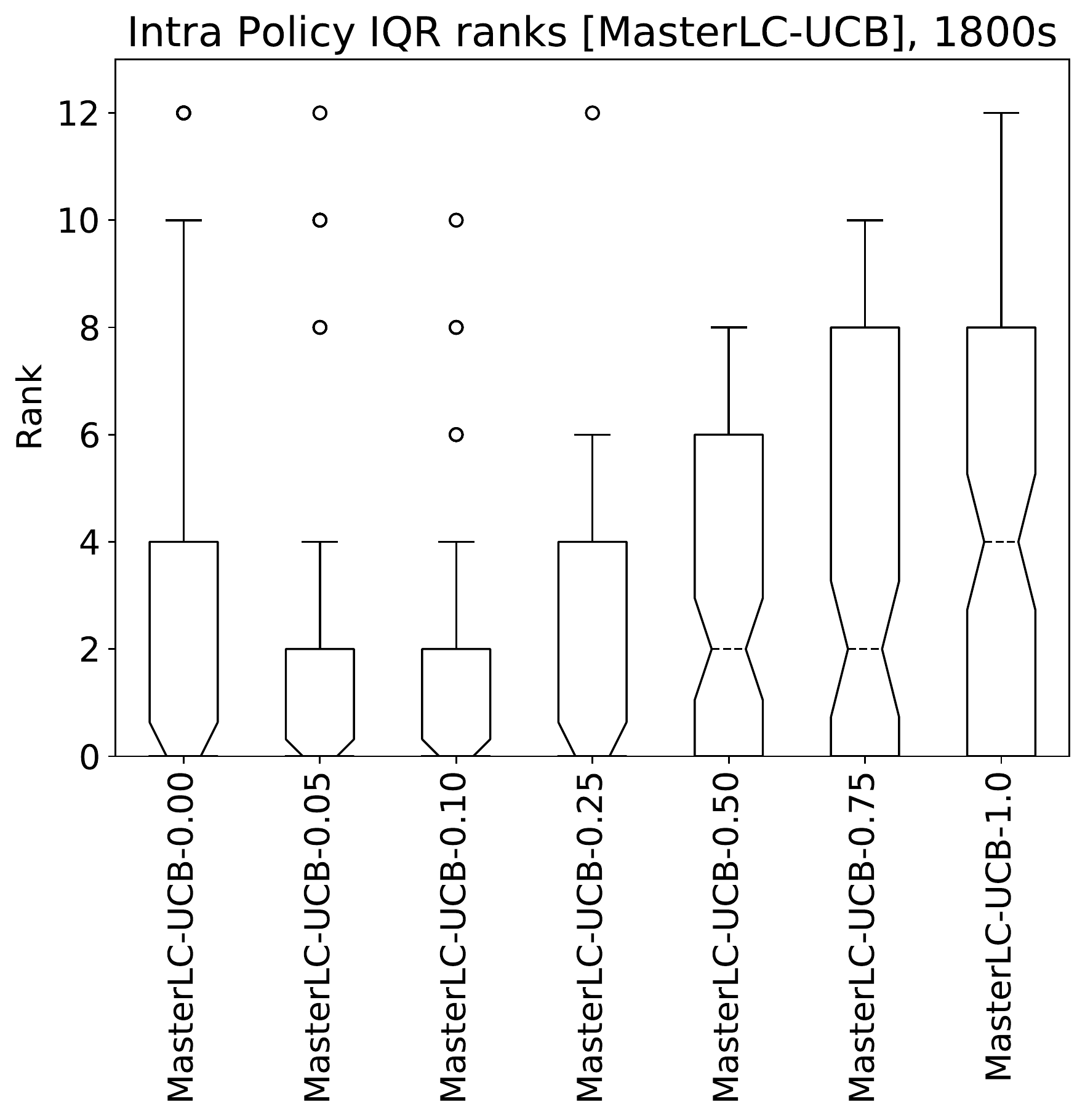}
	\subcaption{Budget 30 minutes.\label{fig:b1800}}
\end{minipage}
\hfill
\begin{minipage}{.22\textwidth}
	\centering
	\includegraphics[width=\textwidth]{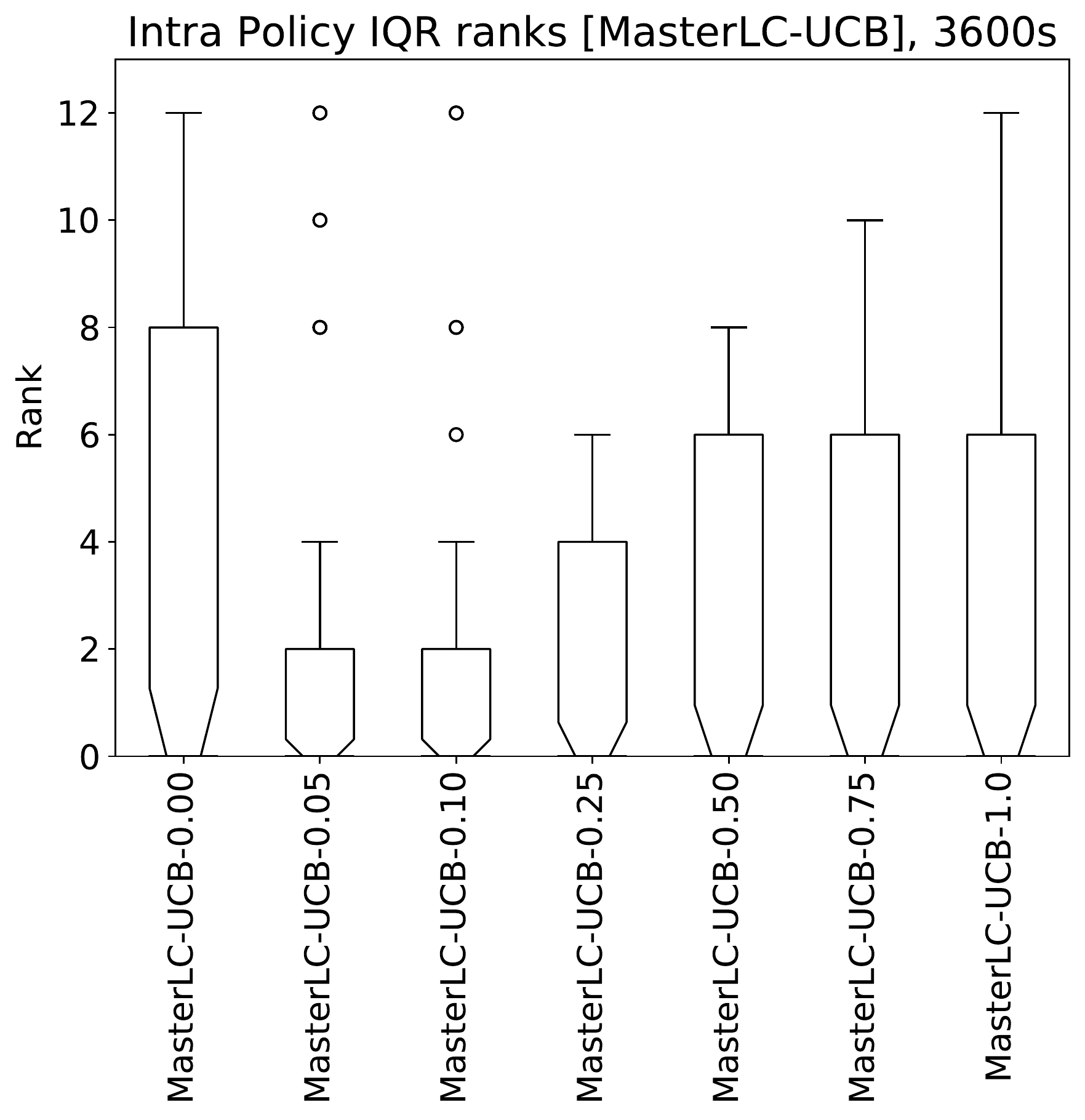}
	\subcaption{Budget 1 hour.\label{fig:b3600}}
\end{minipage}

\caption{Experiment 1: Boxplots of HAMLET Variant 3 ranks. With smaller budgets the results do not change qualitatively.}
	\label{fig:exp1_HAMLET-V3}
\end{figure*}

\begin{figure*}[!t]
\begin{minipage}{.22\textwidth}
	\centering
	\includegraphics[width=\textwidth]{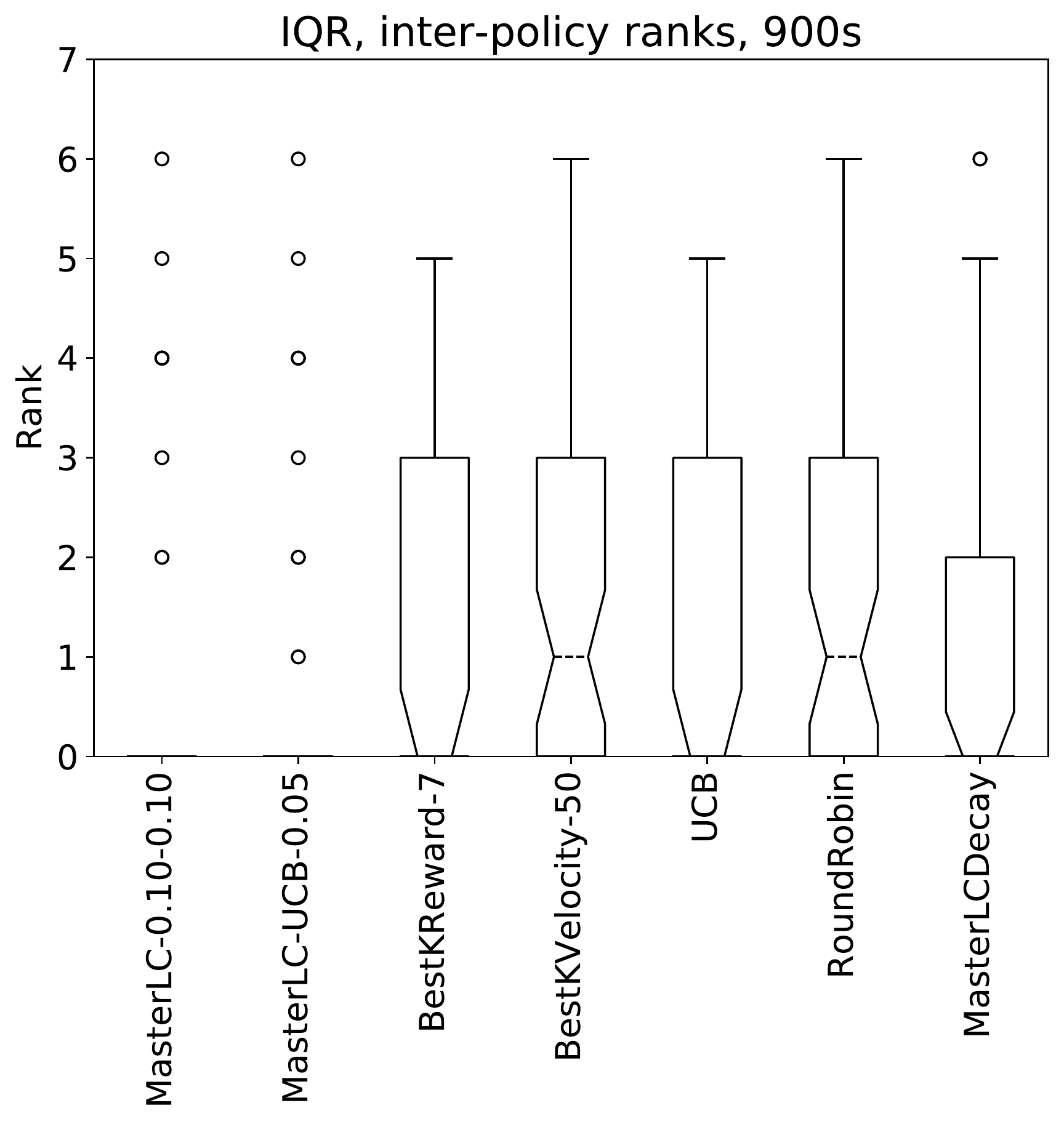}
	\subcaption{\label{fig:b900}Budget 15 minutes.}
\end{minipage}
\hfill
\begin{minipage}{.22\textwidth}
	\centering
	\includegraphics[width=\textwidth]{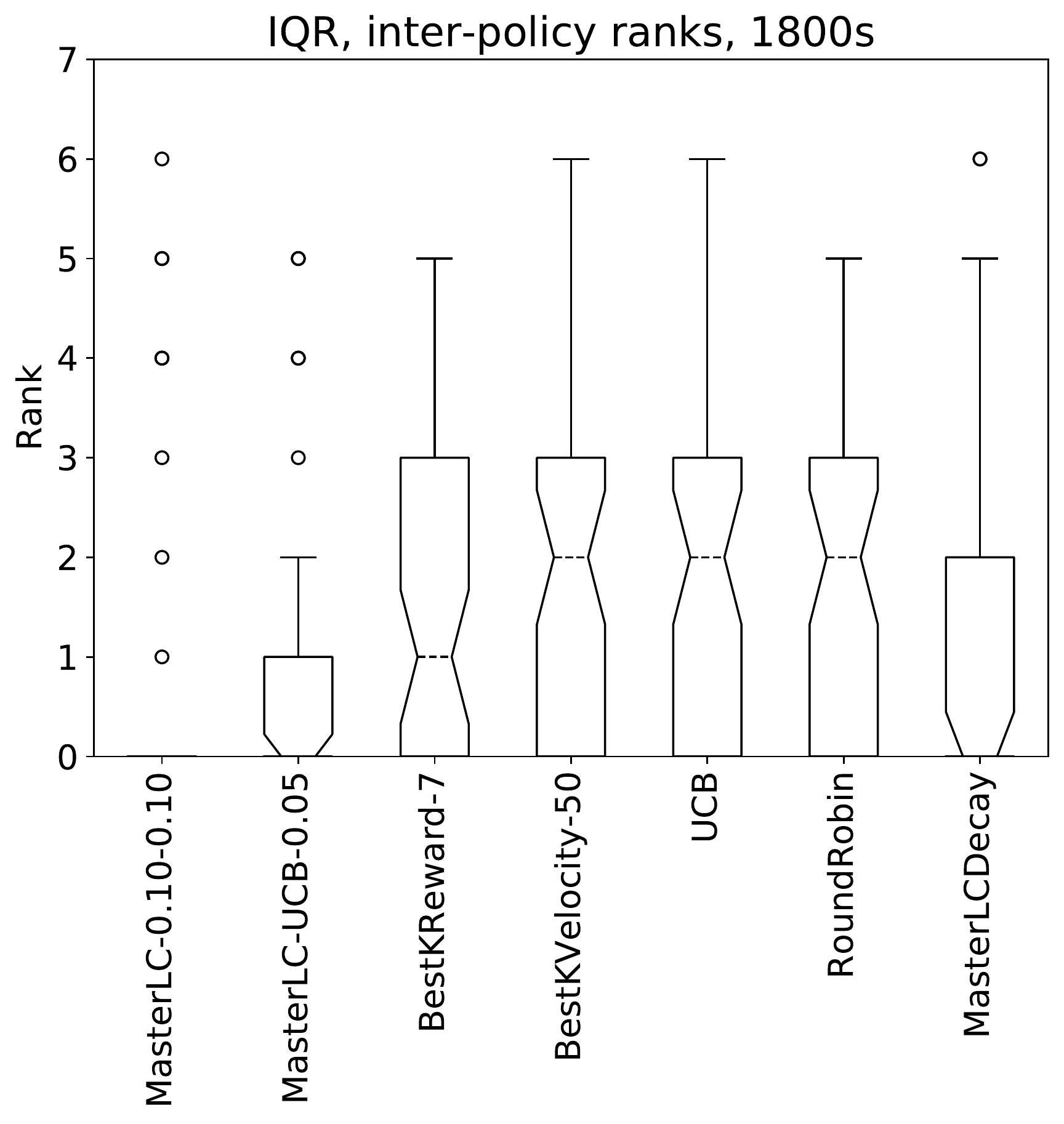}
	\subcaption{Budget 30 minutes.\label{fig:b1800}}
\end{minipage}
\hfill
\begin{minipage}{.22\textwidth}
	\centering
	\includegraphics[width=\textwidth]{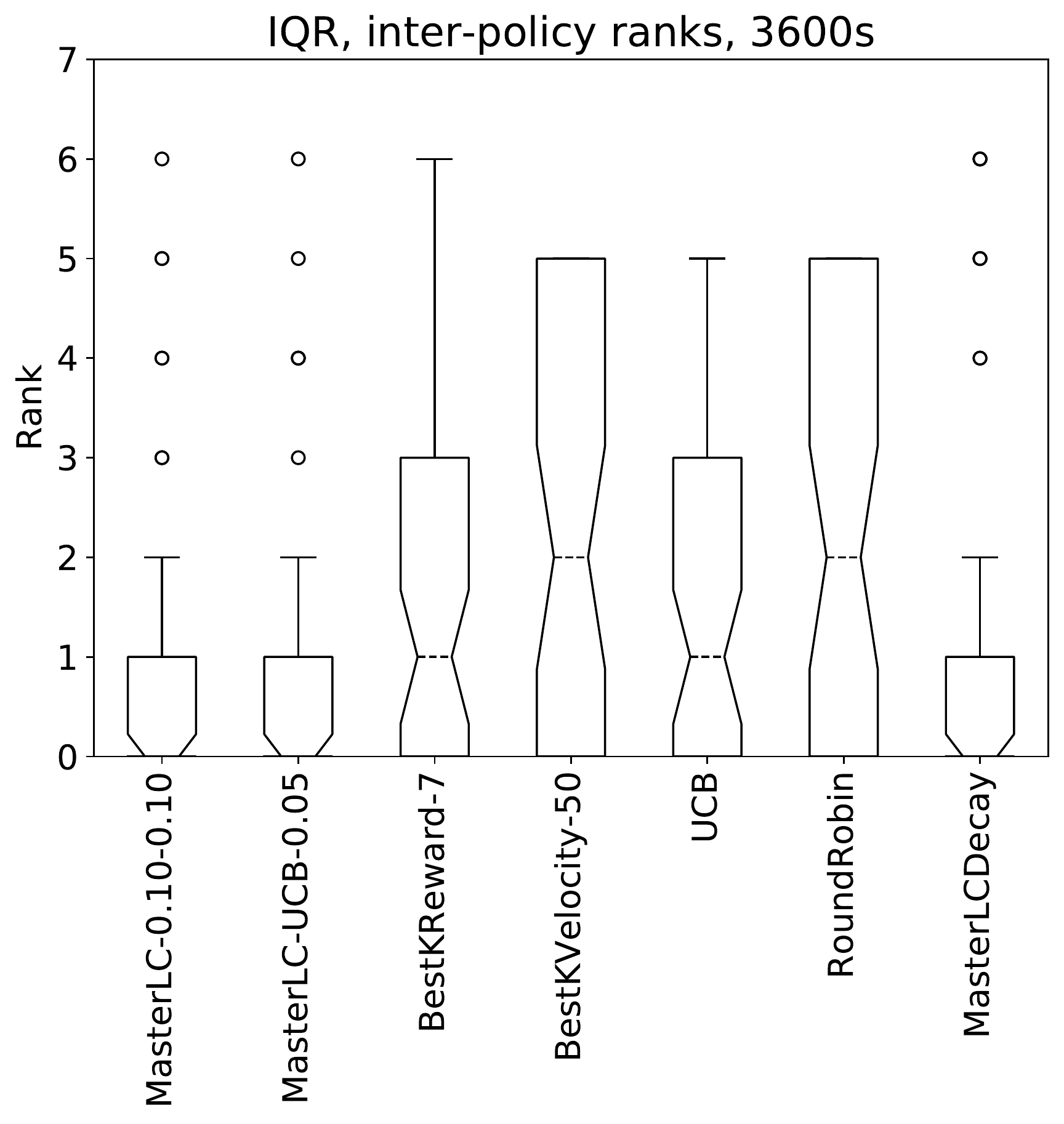}
	\subcaption{Budget 1 hour.\label{fig:b3600}}
\end{minipage}

\caption{Experiment 1: Selected boxplots of ranks for inter-policy comparisons. For $B <  \SI{900}{\second}$  the policies are indistuingishable. }
	\label{fig:exp1}
\end{figure*}

Fig.~\ref{fig:exp1_HAMLET-V1} shows boxplots of HAMLET Variant 1 ranks for different values of $\epsilon_1$ and $\epsilon_2$ for selected  budgets. It confirms the intuition that substantial levels of constant stochasticity, as well as too small levels of stochastic exploration, are detrimental for the performance of Variant 1. We select $\epsilon_1=0.1$ and $\epsilon_2=0.1$ among the parametrizations %performing equally 
for comparing with other bandit policies. 
Fig.~\ref{fig:exp1_HAMLET-V3} shows boxplots of HAMLET Variant 3 ranks for different values of $\rho$. It confirms the intuition that medium to large $\rho$ for scaling the UCB exploration bonus is detrimental for the performance of Variant 3 - as is deactivating the UCB bonus altogether. We select $\rho=0.05$ for comparing with other policies. Fig.~\ref{fig:exp1} compares the ranks of the various bandits for small datasets across different budgets. In particular, HAMLET Variants 1 and 3 achieve favorable performances for all depicted $B$.

\begin{figure*}[!t]
\begin{minipage}{.32\textwidth}
	\centering
	\includegraphics[width=\textwidth]{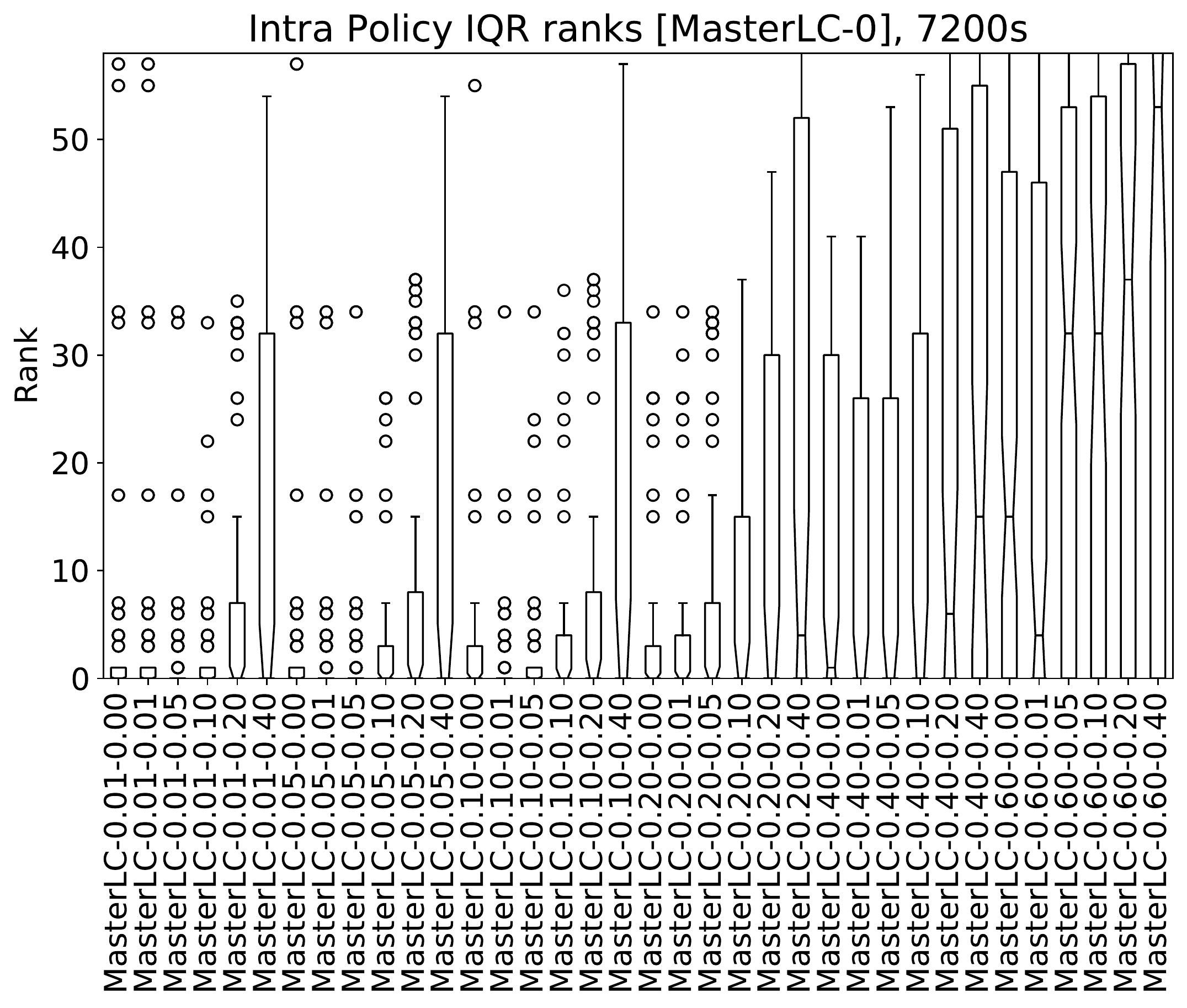}
	\subcaption{Budget 2 hours.}
\end{minipage}
\hfill
\begin{minipage}{.32\textwidth}
	\centering
	\includegraphics[width=\textwidth]{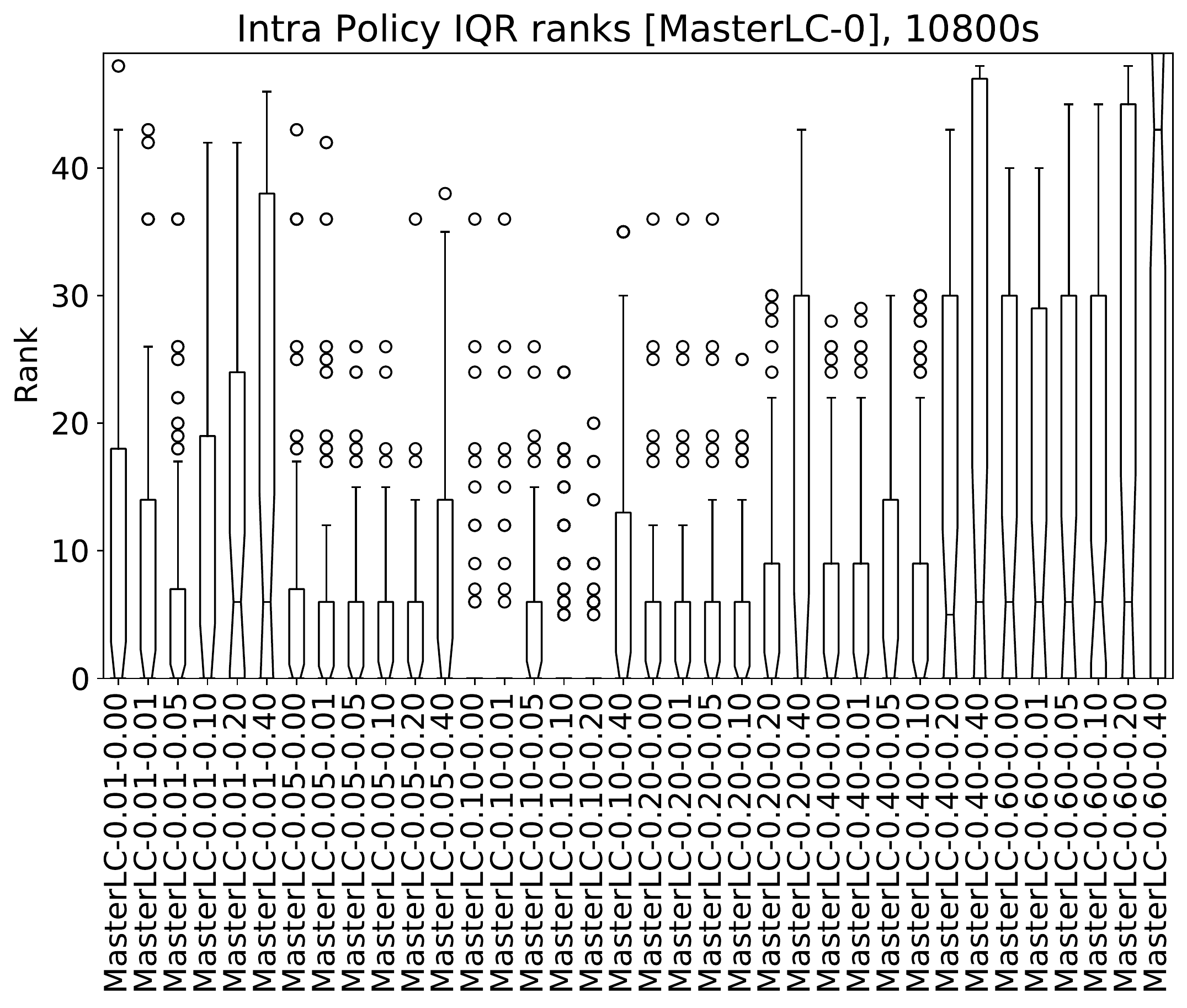}
	\subcaption{Budget 3 hours.}
\end{minipage}
\hfill
\begin{minipage}{.32\textwidth}
	\centering
	\includegraphics[width=\textwidth]{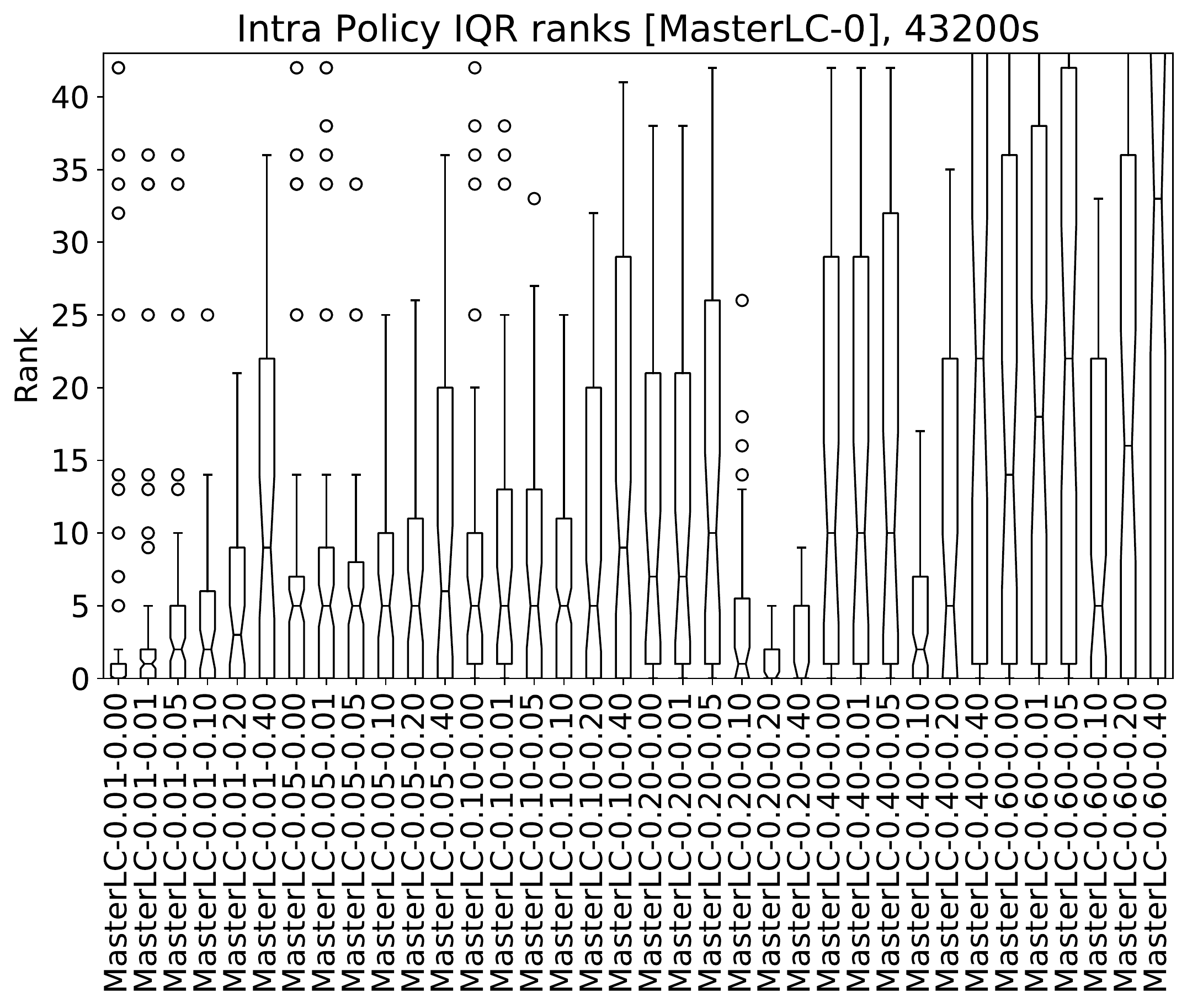}
	\subcaption{Budget 12 hours.}
\end{minipage}

\caption{Experiment 2: Selected boxplots of HAMLET Variant 1 ranks. }
	\label{fig:exp2_HAMLET-V1}
\end{figure*}

\begin{figure*}[!t]
\begin{minipage}{.2\textwidth}
  \centering
  \includegraphics[width=\textwidth]{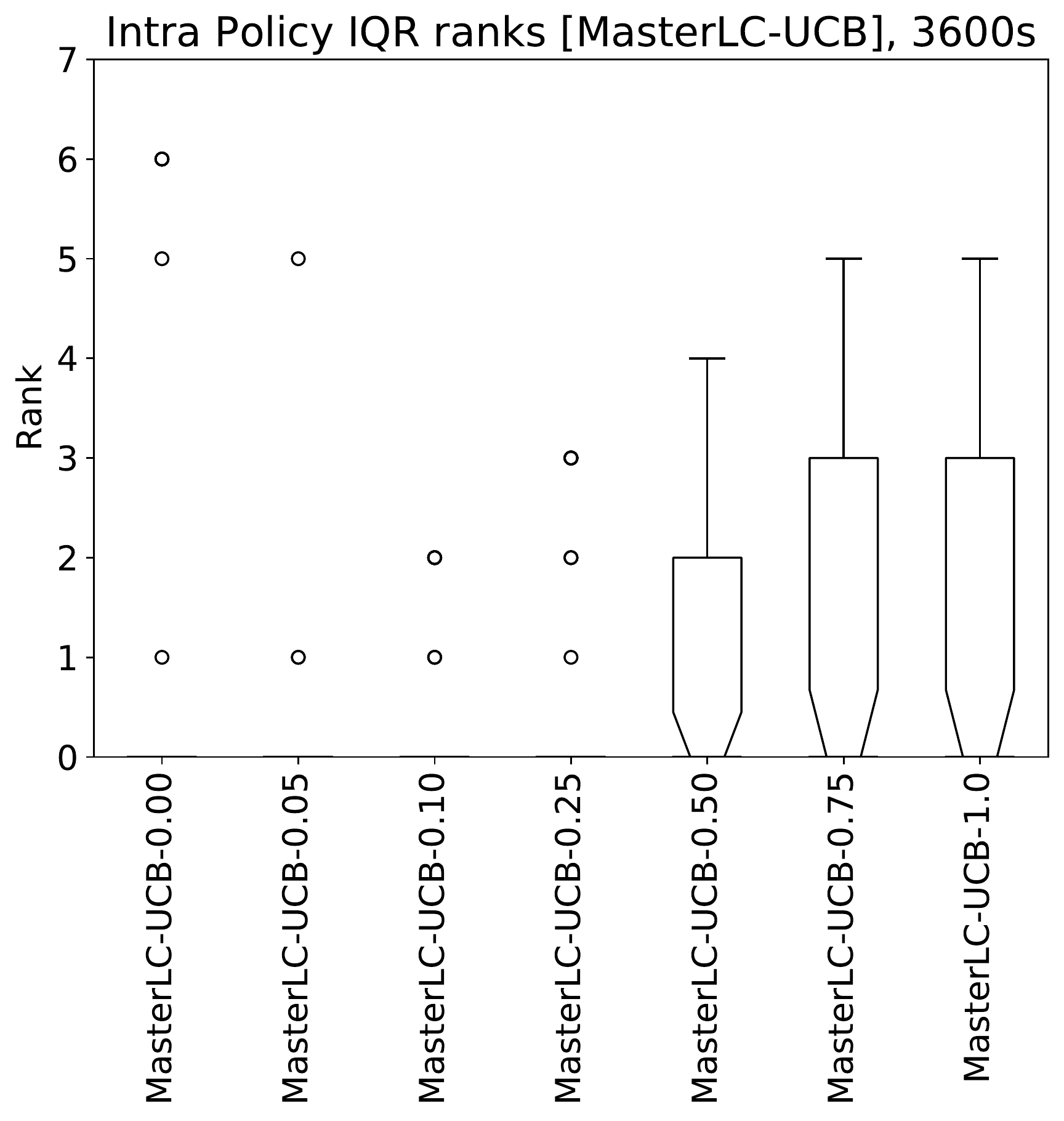}
  \subcaption{Budget 1 hour. }
\end{minipage}
\hfill
\begin{minipage}{.2\textwidth}
	\centering
	\includegraphics[width=\textwidth]{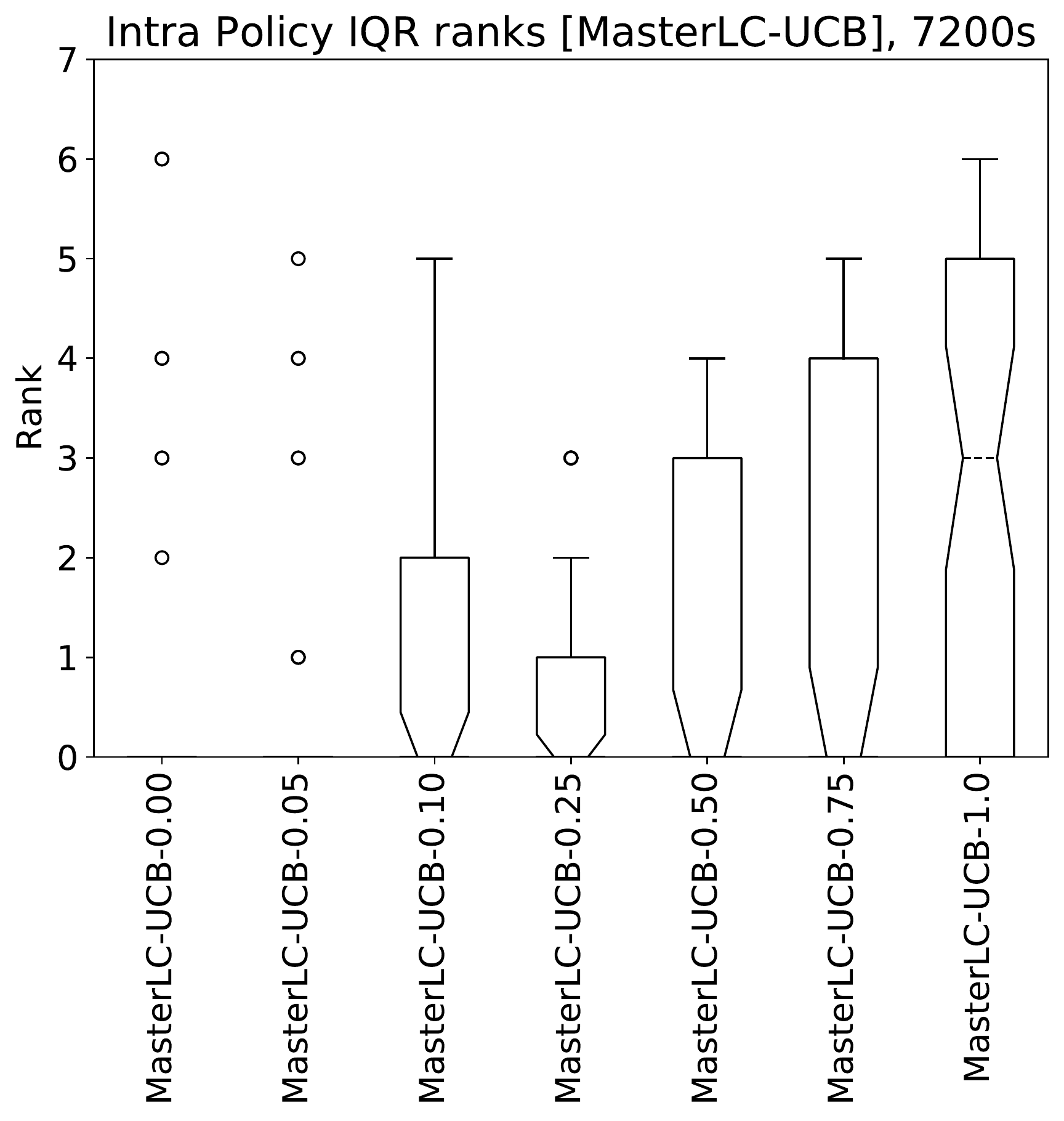}
	\subcaption{Budget 2 hours.}
\end{minipage}
\hfill
\begin{minipage}{.2\textwidth}
	\centering
	\includegraphics[width=\textwidth]{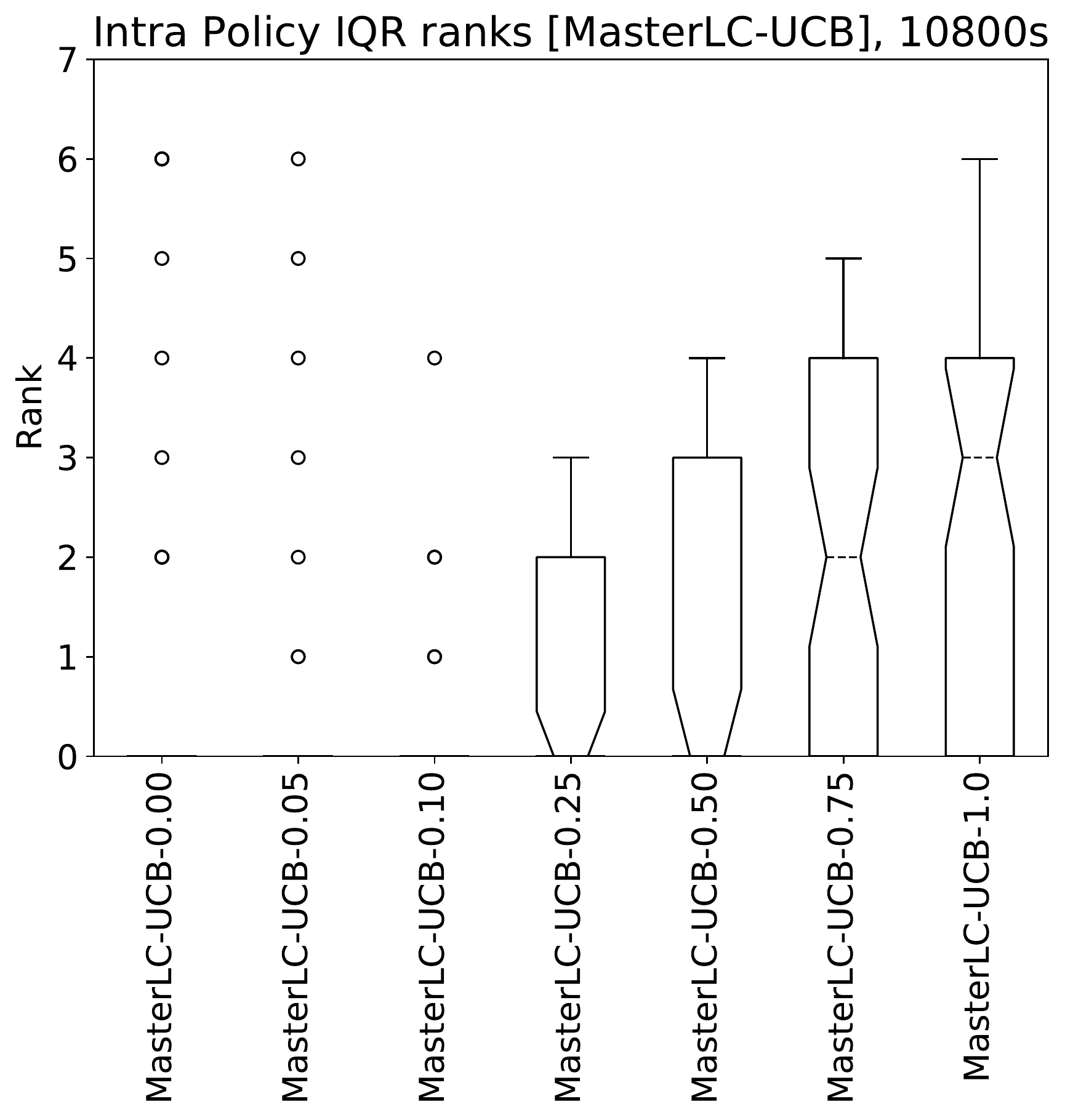}
	\subcaption{Budget 3 hours.}
\end{minipage}
\hfill
\begin{minipage}{.2\textwidth}
	\centering
	\includegraphics[width=\textwidth]{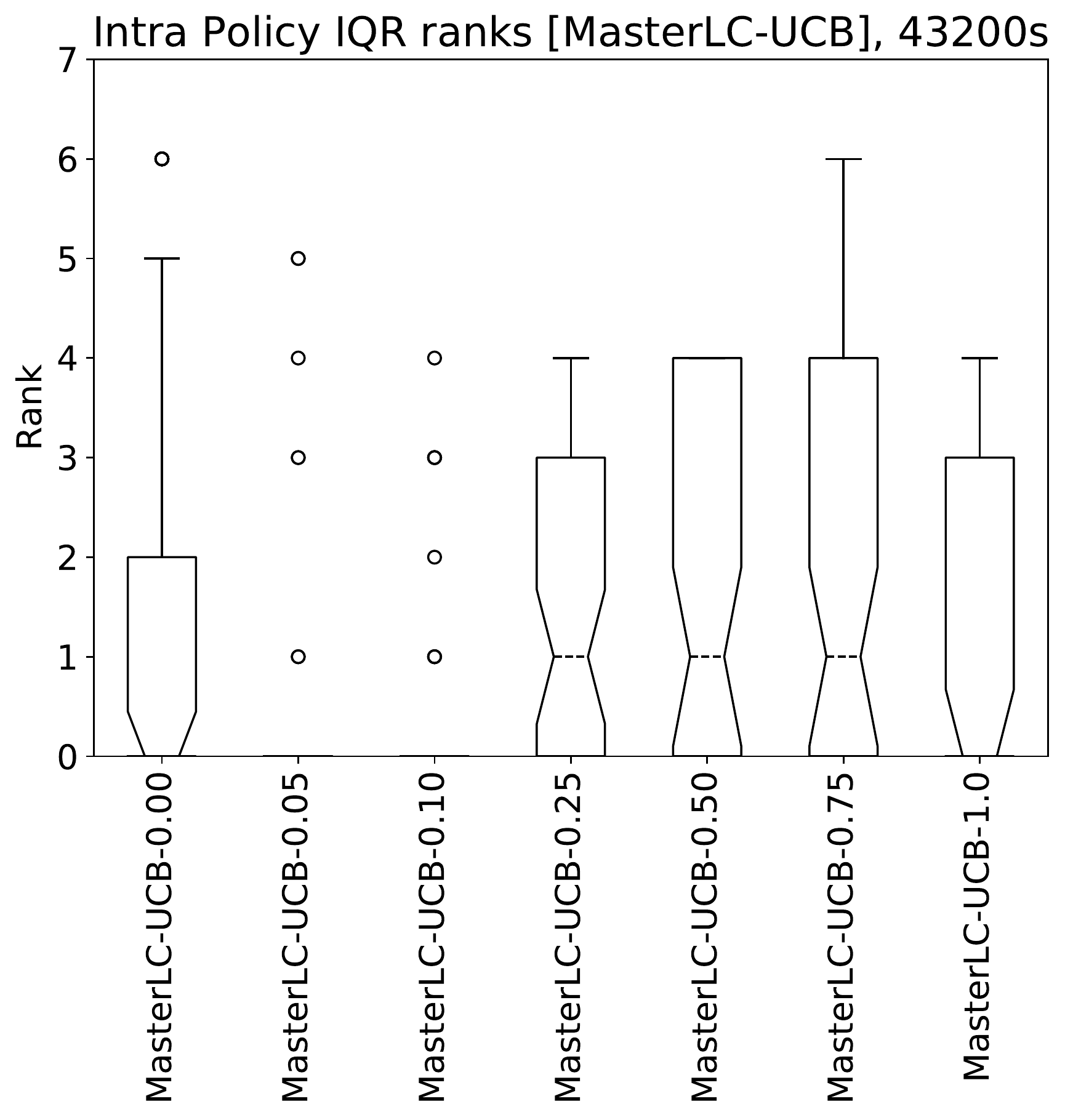}
	\subcaption{Budget 12 hours.}
\end{minipage}

\caption{Experiment 2: Selected boxplots of HAMLET Variant 3 ranks. }
	\label{fig:exp2_HAMLET-V3}
\end{figure*}

\begin{figure*}[!t]
\begin{minipage}{.2\textwidth}
  \centering
  \includegraphics[width=\textwidth]{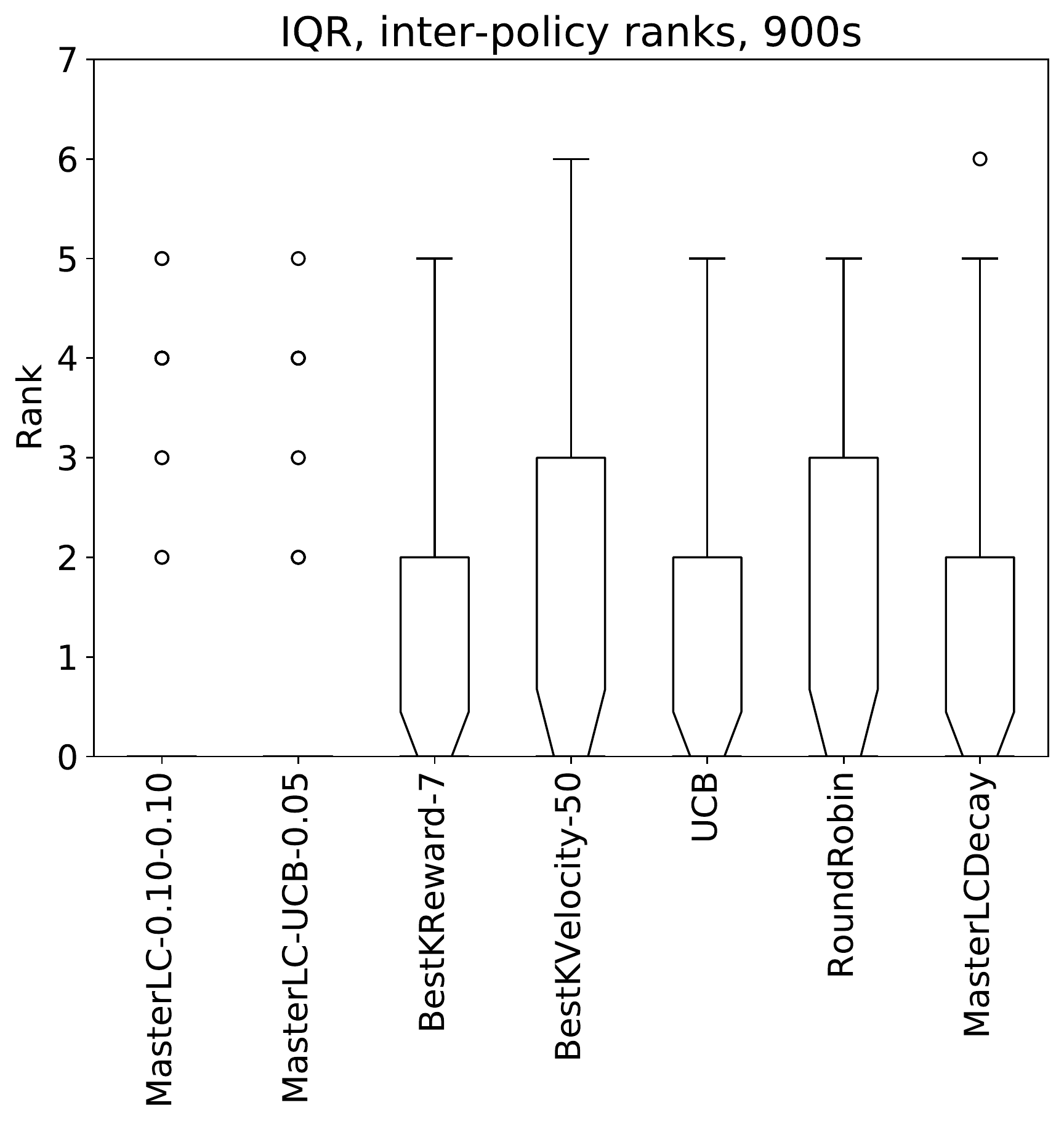}
  \subcaption{Budget 15 minutes. }
\end{minipage}
\hfill
\begin{minipage}{.2\textwidth}
	\centering
	\includegraphics[width=\textwidth]{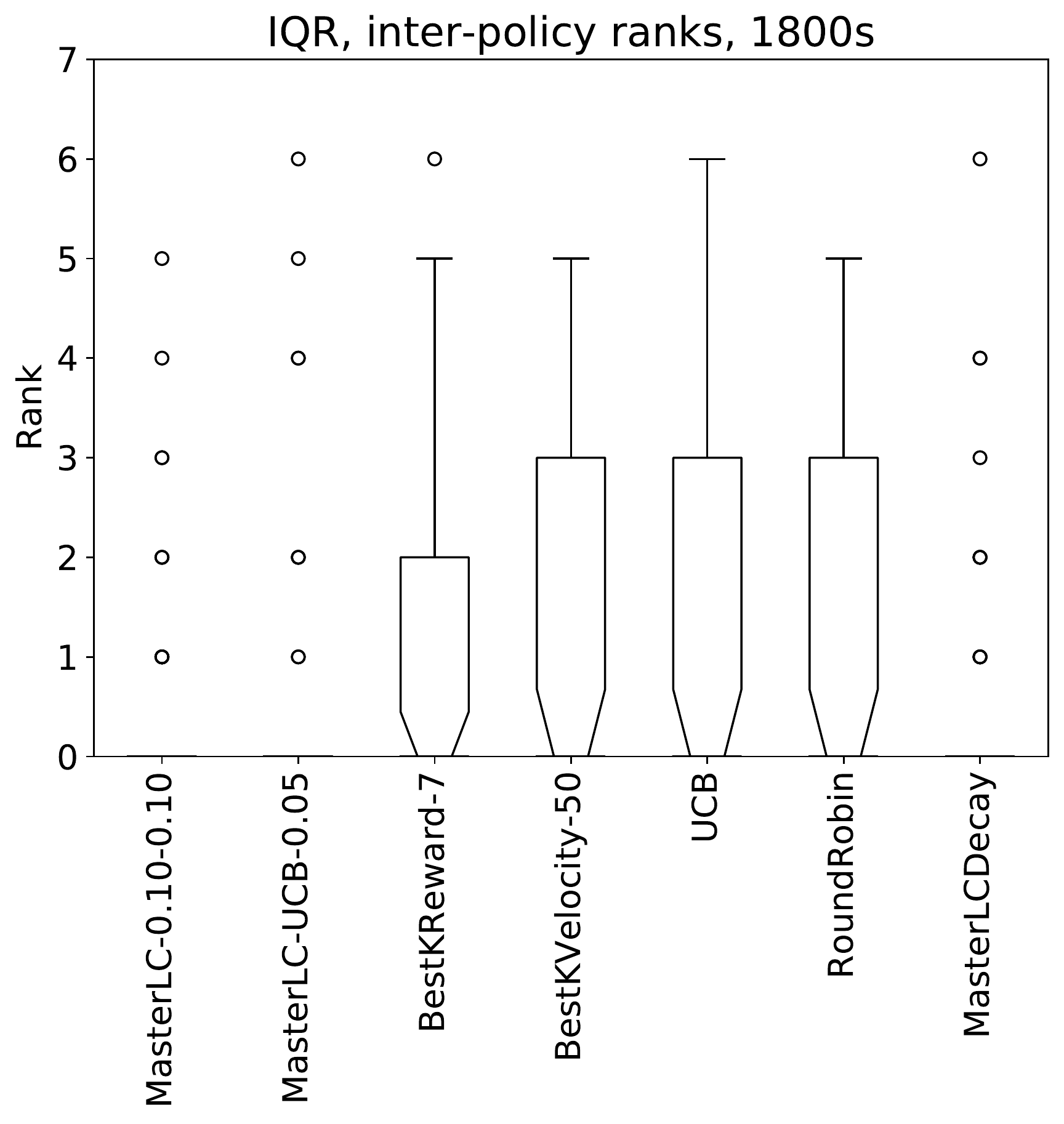}
	\subcaption{Budget 30 minutes.}
\end{minipage}
\hfill
\begin{minipage}{.2\textwidth}
	\centering
	\includegraphics[width=\textwidth]{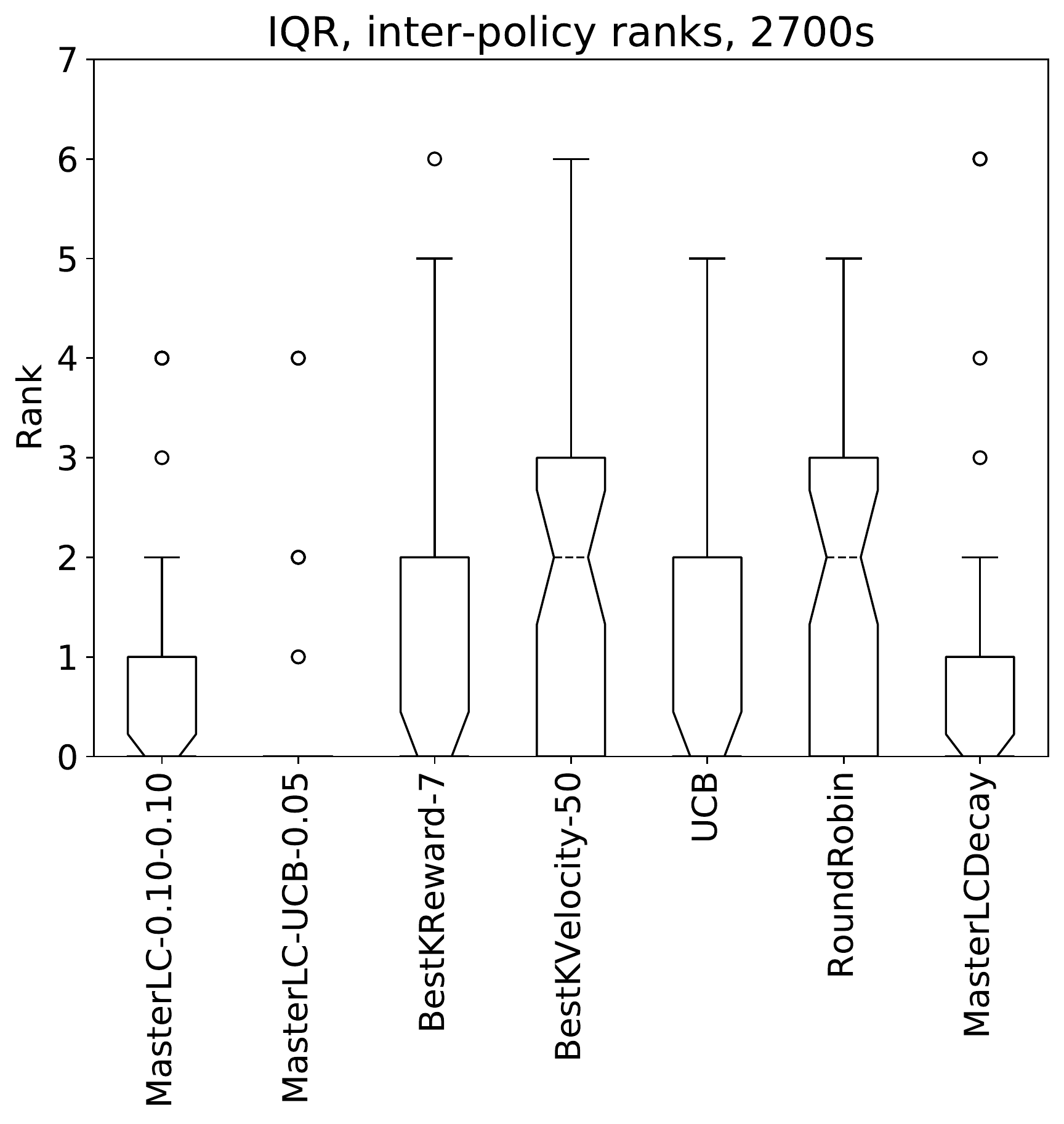}
	\subcaption{Budget 45 minutes.}
\end{minipage}
\hfill
\begin{minipage}{.2\textwidth}
	\centering
	\includegraphics[width=\textwidth]{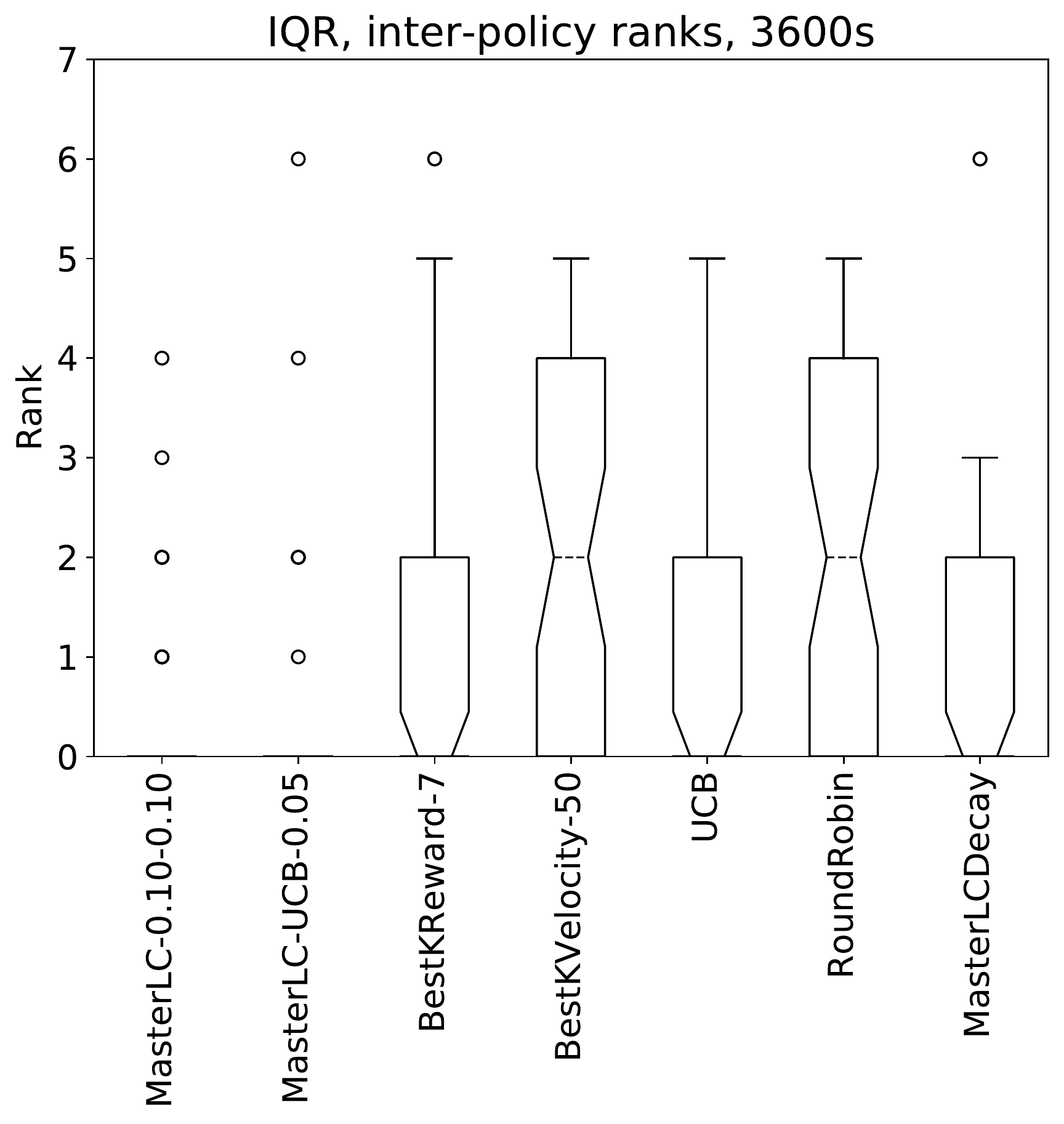}
	\subcaption{Budget 1 hour.}
\end{minipage}

\begin{minipage}{.2\textwidth}
  \centering
  \includegraphics[width=\textwidth]{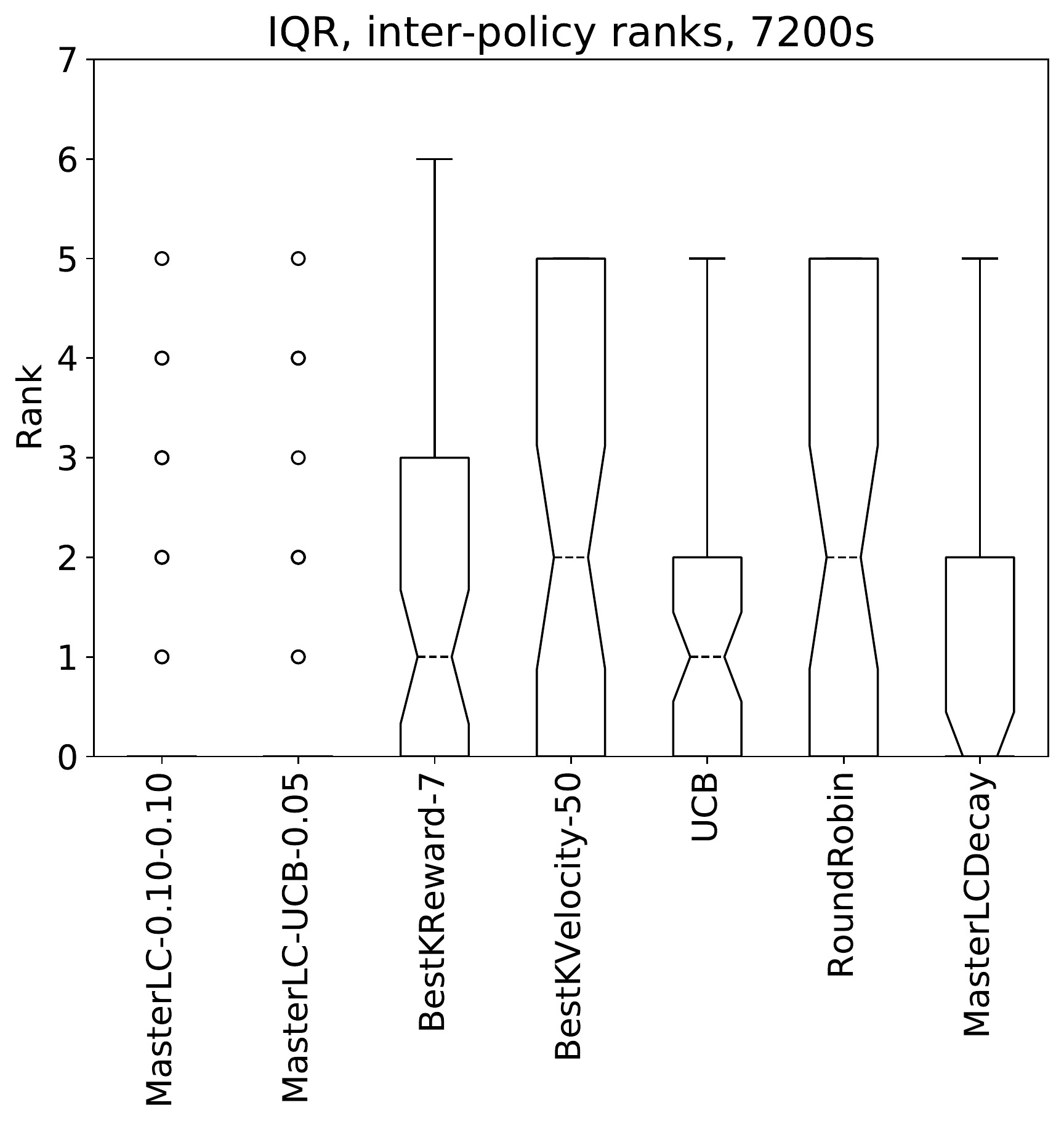}
  \subcaption{Budget 2 hours. }
\end{minipage}
\hfill
\begin{minipage}{.2\textwidth}
	\centering
	\includegraphics[width=\textwidth]{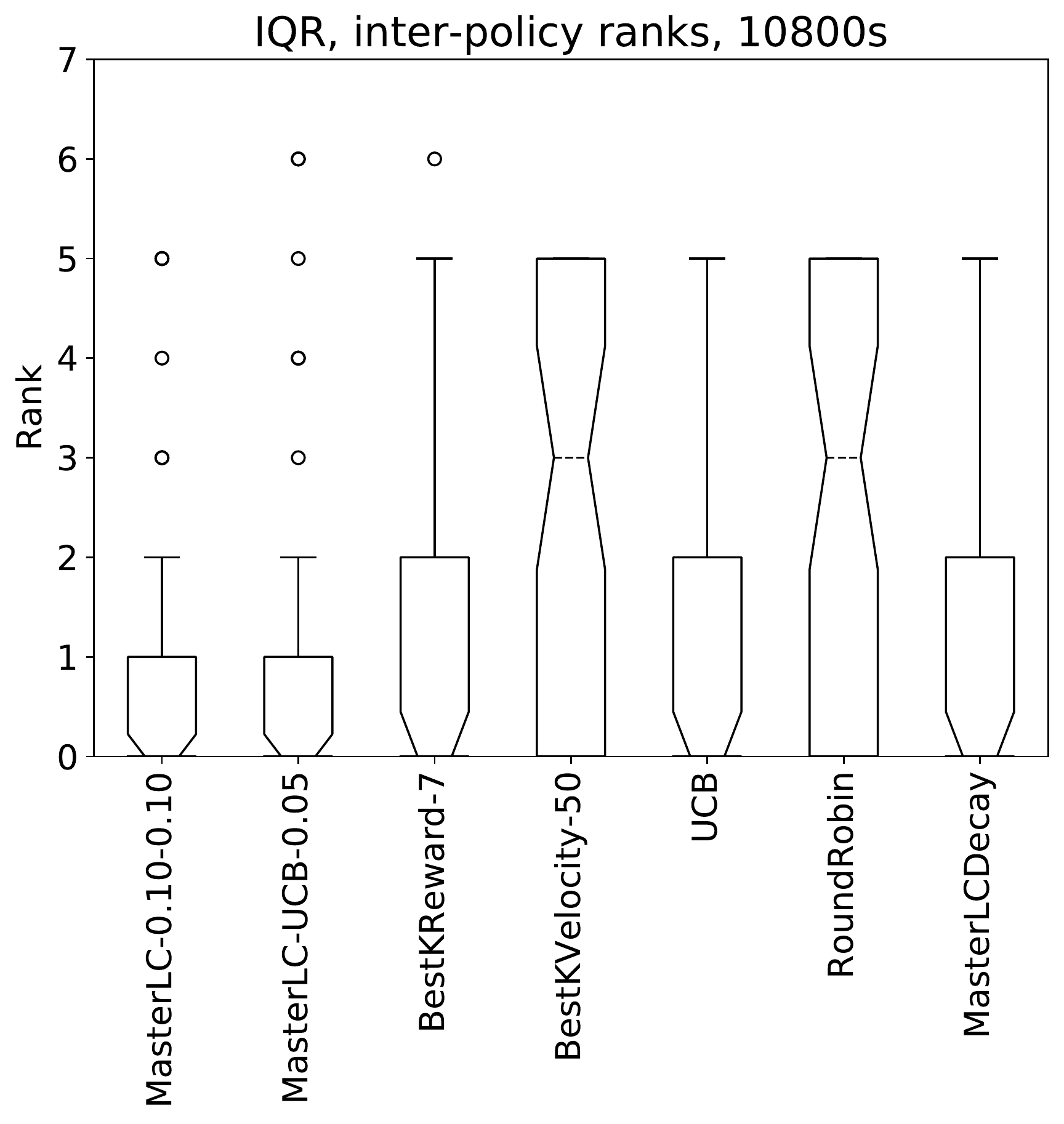}
	\subcaption{Budget 3 hours.}
\end{minipage}
\hfill
\begin{minipage}{.2\textwidth}
	\centering
	\includegraphics[width=\textwidth]{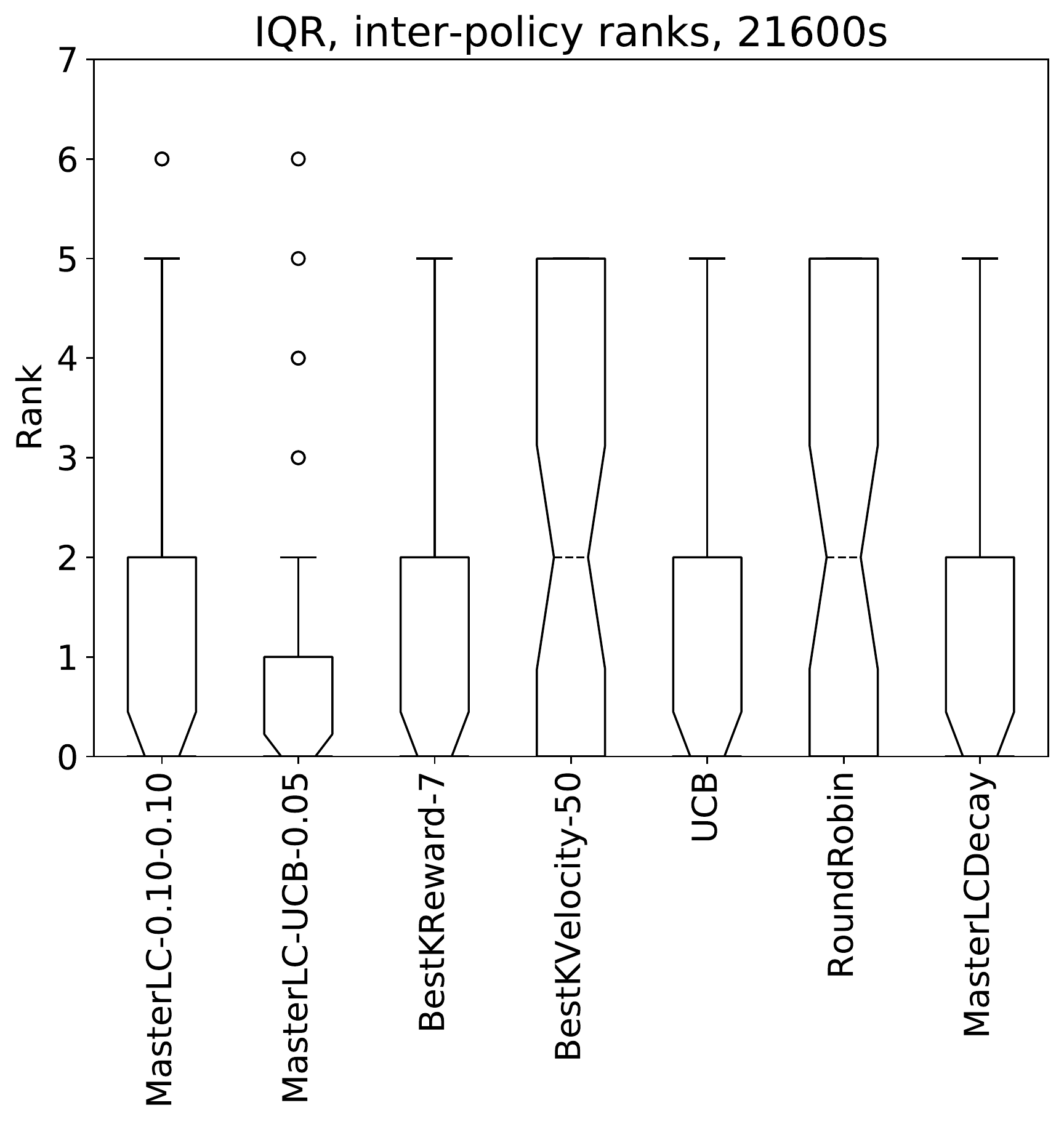}
	\subcaption{Budget 6 hours.}
\end{minipage}
\hfill
\begin{minipage}{.2\textwidth}
	\centering
	\includegraphics[width=\textwidth]{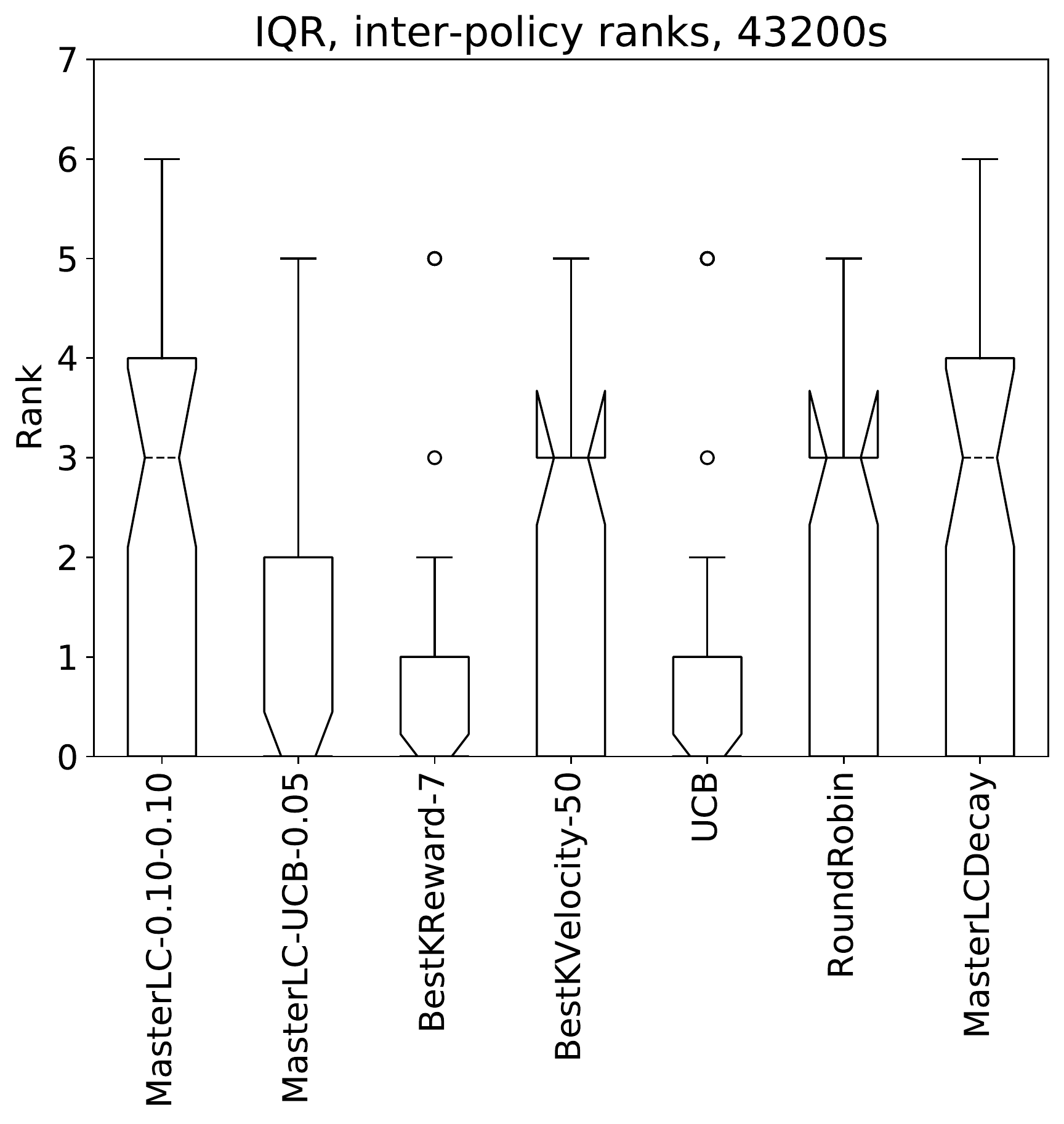}
	\subcaption{Budget 12 hours.}\label{fig:exp2_exception}
\end{minipage}

\caption{Experiment 2: Boxplots of ranks for inter-policy comparisons.  }
	\label{fig:exp2}
\end{figure*}

%\subsection{Experiment 2: Bigger Datasets}
Similar to Fig.~\ref{fig:exp1_HAMLET-V1}, Fig.~\ref{fig:exp2_HAMLET-V1} shows selected boxplots of HAMLET Variant 1 ranks for different values of $\epsilon_1$ and $\epsilon_2$ across a range of budgets. Fig.~\ref{fig:exp2_HAMLET-V1} re-confirms the intuition that high levels of constant stochasticity, as well as too small levels of stochastic exploration, are detrimental for the performance of Variant 1. We confirm the selection of $\epsilon_1=0.1$ and $\epsilon_2=0.1$. Similar to Fig.~\ref{fig:exp1_HAMLET-V3}, Fig.~\ref{fig:exp2_HAMLET-V3} shows boxplots of HAMLET Variant 3 performance. It confirms the intuition that a medium to large $\rho$ value is detrimental for the performance of Variant 3. We confirm $\rho=0.05$. Fig.~\ref{fig:exp2} illustrates the boxplot of ranks for bigger datasets for all eight budgets. HAMLET Variants 1 (\textit{MasterLC}-$\epsilon_1$-$\epsilon_2$) and 3 (\textit{MasterLC-UCB-$\rho$}) achieve a favorable performance. At higher budgets, \textit{BestKRewards-7} and \textit{UCB} improve in their rankings relative to the HAMLET policies and can even overtake  HAMLET Variant 3 for the 12-hour budget. 

\section{Discussion}\label{sec:discussion}
Section \ref{sec:experiments} indicates that HAMLET Variants 1 and 3 benefit from the time-awareness and the learning curve extrapolation capability. Exploration needs to be encouraged in a moderate manner. Too high levels of stochasticity or exploration bonus reduce the algorithm selection performance (Fig.~\ref{fig:exp1_HAMLET-V1}, \ref{fig:exp1_HAMLET-V3}, \ref{fig:exp2_HAMLET-V1}, \ref{fig:exp2_HAMLET-V3}). HAMLET Variant 2 performs worst of the HAMLET variants. For several low to medium budgets in each experiment, HAMLET Variants 1 and 3 perform better than the competitor policies (Fig.~\ref{fig:exp1}, \ref{fig:exp2}). For budgets outside of that range, HAMLET Variants 1 and 3 perform mostly on par with the state of the art BestK-Reward policy - except for the largest budget (Fig.~\ref{fig:exp2_exception}), where BestK-Rewards and UCB1 achieve better results than even HAMLET Variant 3. 

\begin{figure}[!t]
\centering
	\includegraphics[width=0.67\columnwidth]{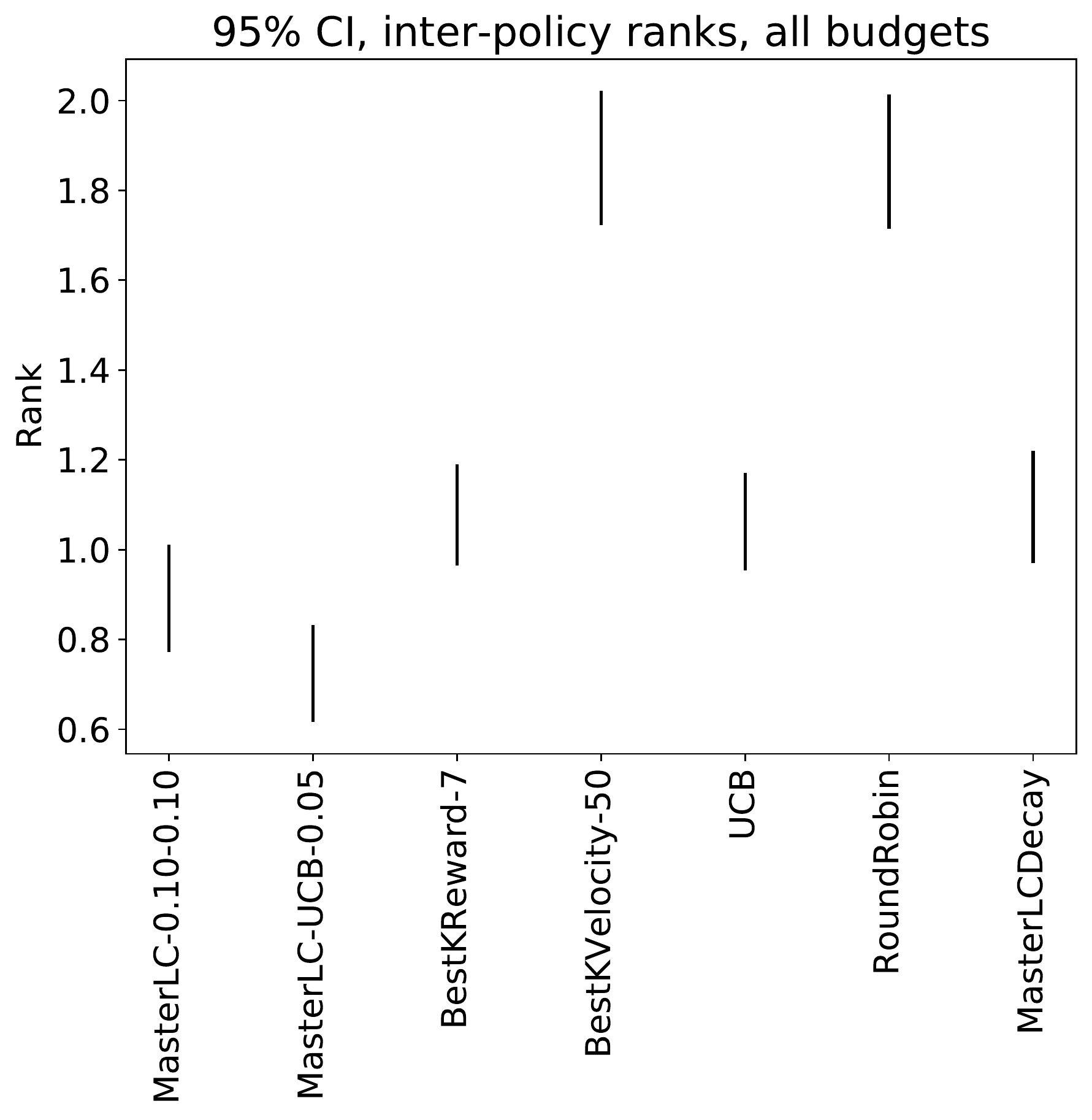}
%\caption{Intervals of 95\% confidence of the different policies' mean ranks, aggregated over all datasets and budgets. %Note that HAMLET Variant 3 has a non-overlapping confidence interval with all other policies except for Variant 1.
\caption{Confidence intervals for policies' mean ranks.
}\label{fig:totalCI}
\end{figure}

The boxplots in Fig.~\ref{fig:exp1} and \ref{fig:exp2} suggest that HAMLET performs better than the compared-to policies for a range of budgets. However, we did not find on a per budget level a consistent trend that is statistically significant at the 95\% level in the differences of the policies' mean ranks. Fig.~\ref{fig:totalCI} shows the aggregation of all 1,485 runs (99 traces $\times$ 15 budget levels). Here, HAMLET Variant 3 - learning curve extrapolation combined with an uncertainty bonus for exploration - achieves a better performance than the other policies (except the HAMLET Variant 1). Because the respective confidence intervals do not overlap, that better performance is statistically significant at the 95\% level. 
Therefore, we conclude that the experiments - in particular, the performance of HAMLET Variant 3 - confirm the hypothesis introduced in Section \ref{sec:intro}. Considering that the applied technique is straightforward (Eq.~\ref{eq:atan}), we argue that these results encourage future work on integrating more sophisticated learning curve techniques into multi-armed bandits for algorithm selection. Advanced approaches should only increase the relative advantage of HAMLET over alternative approaches.

In an evaluation not presented in this work for page limitations, we also verified that the trends of the results do not change when we compare for each budget the different policy groups' best-performing policies against each other. We chose to focus the analysis on the results recorded in Section \ref{sec:experiments} as it is the more realistic setting from an end-user perspective.

Finally, we observed during the experiments that the BestK-Velocity policy usually performs much worse than the BestK-Rewards strategy or the UCB strategy. 

Traditional multi-armed bandit algorithms can rely on incremental updates of running averages of the observed rewards with complexity $\mathcal{O}(1)$ for action selection. Each iteration lasting $\Delta t$ and observing $n$ test scores, the Best-K algorithms require comparing $n$ novel observations with the list of $k$ top observations, implying a complexity of $\mathcal{O}(n)$.  In contrast, HAMLET filters in each iteration the $n$ new observations for monotonicity, which is also $\mathcal{O}(n)$. Besides, HAMLET incurs the complexity of curve-fitting the resulting $m$ monotonically increasing observations found so far using \cite{virtanen2020scipy}. Depending on the used solver algorithm that might involve, e.g.~cubically scaling matrix inversions, i.e.~$\mathcal{O}(m^3)$.%\footnote{In the experiments, scipy required little computation time to fit Eq.~\ref{eq:atan} when compared to the iteration interval of $\Delta t = 10s$: usually less than 200 ms.}.
%https://www.freecodecamp.org/news/data-science-with-python-8-ways-to-do-linear-regression-and-measure-their-speed-b5577d75f8b/

Note that this work relies on hyperparameter tuning traces of \cite{8851978} to evaluate the different bandits for solving the algorithm selection problem. These traces are limited to classification datasets. To move to other learning problems, e.g.~regression, the learning curve model in Eq.~\ref{eq:atan} requires adaptation. 

\section{Conclusion and Future Work}\label{sec:conclusion}
This work introduced HAMLET, a multi-armed bandit for algorithm selection that is able to account for the progress of time and is capable of extrapolating learning curves. Experiments with a range of bandit policy parametrizations show that even a straightforward approach to extrapolate learning curves is a valuable amendment for the bandit-based algorithm selection problem. Statistical analysis shows that the HAMLET Variants 1-3 are at least as good as standard bandit approaches. Notably, HAMLET Variant 3, which combines learning curve extrapolation with a scaled UCB exploration bonus, performs superior to all non-HAMLET variants (Fig.~\ref{fig:totalCI}). 

This work motivates multiple areas of future work. First, more sophisticated learning curve modeling approaches, e.g.~the BNN-based learning curve predictors in \cite{Klein2017LearningCP}, lend themselves for investigation. Another avenue to further improve performance is to integrate meta-learning concepts, e.g.~by evolving HAMLET into a contextual bandit.  Third, we plan to adapt Eq.~\ref{eq:atan} to regression problems.

\section*{Acknowledgment}
We thank Tobias Jacobs for providing valuable discussions during HAMLET's design.

\bibliography{ijcnn2020-refs}
\bibliographystyle{IEEEtran} 

\end{document}